\documentclass{article} % For LaTeX2e
\usepackage{iclr2027_conference,times}

\usepackage{amsmath,amsfonts,bm}

\def\eqref#1{equation~\ref{#1}}
\def\1{\bm{1}}

\DeclareMathAlphabet{\mathsfit}{\encodingdefault}{\sfdefault}{m}{sl}
\SetMathAlphabet{\mathsfit}{bold}{\encodingdefault}{\sfdefault}{bx}{n}

\usepackage{hyperref}
\usepackage{url}
\usepackage{booktabs}       % professional-quality tables
\usepackage{amsfonts}       % blackboard math symbols
\usepackage{nicefrac}% compact symbols for 1/2, etc.
\usepackage{graphicx}
\usepackage{microtype}      % microtypography
\usepackage{xcolor}         % colors
\usepackage{floatrow}
\usepackage{multirow}

\title{The Search for Statistical Indistinguishability Between Real and Fake Images}
\iclrfinalcopy

\author{Mohaimin Rafi*~~~~~Otavio Rosa*~~~~~Anshuman Chhabra~~~~~Utkarsh Ojha  \\\\
University of South Florida\\\\
\texttt{\{srafi, omrosa, anshumanc, uojha\}@usf.edu} \\
}

\begin{document}

\maketitle

\def\thefootnote{}\footnotetext{*Equal contribution}

\begin{abstract}
Images produced by generative models appear strikingly similar to real ones. In this work, we study whether perfection has been reached by generated content in terms of matching the properties of real data. We train deep neural networks to separate real data from fake and observe their test accuracies. Only a chance performance has any scope of proving statistical indistinguishability. However, since generative models are known to sometimes introduce small errors, e.g., low level checkerboard artifacts, we create progressive abstractions of the original real/fake images to potentially bring them to a space where those artifacts get wiped away. These include operations such as blurring, downsizing, cropping, converting RGB images into edge/semantic maps, etc. To our surprise, almost none of the abstractions are able to make fake data indistinguishable from real. The only cases when neural network based classifiers fail to separate the two categories is when the data is abstracted to such an extent that the original content seems mostly lost to the human eye, e.g., downsizing a $256 \times 256$ RGB image of human face to $4 \times 4$ resolution. We observe this trend across four different datasets of varying complexity – MNIST, CelebA, FFHQ and GenImage – and across various types of generative models – GANs, VAEs, diffusion models. This shows that the nature of fake artifacts extends all the way from low level to very high level in images. The simple implication is that fake images are much more different from real than we are used to think.
\end{abstract}

\section{Introduction}
The goal of an image generative model is to produce realistic images from a target distribution. Perfection at this task would mean that the generated images are \textit{indistinguishable} from real ones. Recent generative models have seemingly made a lot of progress towards this goal~\citep{midjourney, dall-e2}. Typically, this idea of \textit{indistinguishability} is limited to human judgment, where a human person should not be able to identify which image is real and which one is fake. However, there is an even more challenging test available for these models to pass successfully. The generated images and real images need to be indistinguishable not just to humans, but also to a neural network based classifier trained to distinguish them, i.e., its performance on the binary real-vs-fake classification task should stay at chance-level ($\sim50\%$). This is known as the two-sample test~\citep{two-sample}. In this work, we study the extent to which generative models have come close to passing this ultimate test.

\begin{figure}[t]
    \centering
    \includegraphics[width=0.98\textwidth]{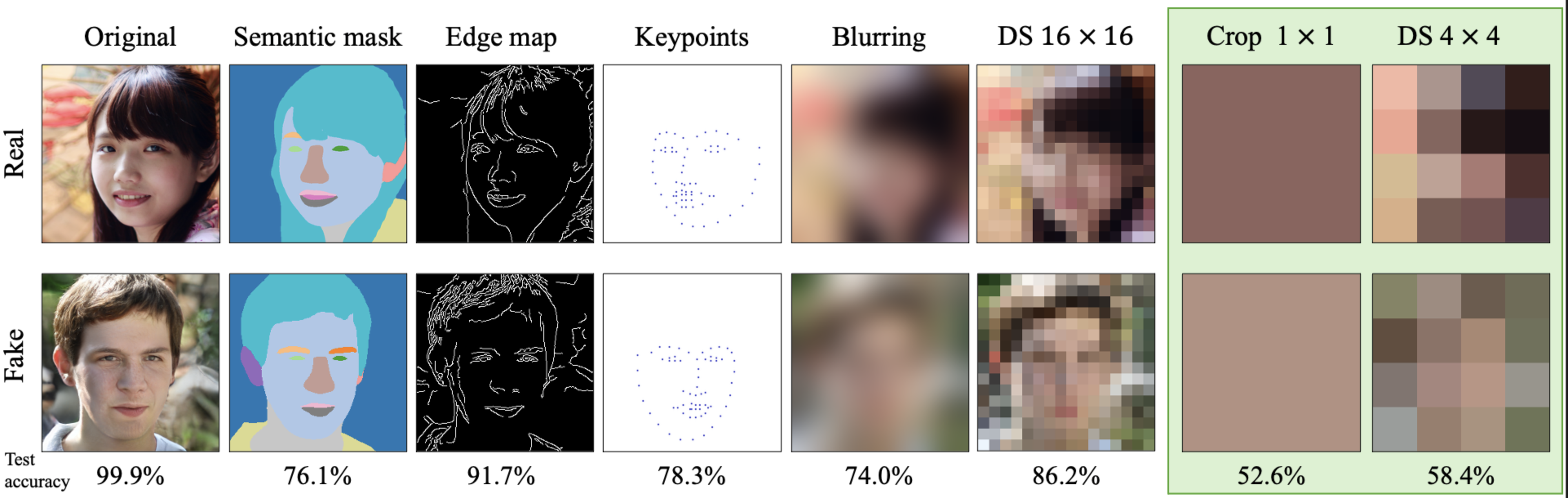}
    \caption{The first column corresponds to a classifier being trained to distinguish clean real  (top) and fake images (bottom), achieving a high test accuracy because of the presence of artifacts. In each subsequent column, we transform the original images with the hope of masking away the artifact signal; e.g., downsizing (DS) to $16 \times 16$ (6th column). When new classifiers are trained for each column, the accuracy still remains high. For true \textit{ statistical indistinguishability} ($\sim50\%$; last two columns), we need to abstract away most of the image content by, e.g., downsizing to $4 \times 4$ resolution.}
    \label{fig:teaser}
\end{figure}

If we consider the real and fake image data in its original RGB space, a classifier can separate them easily. For example, in Fig.~\ref{fig:teaser} column 1, a classifier achieves close to $100\%$ test accuracy in distinguishing real human faces~\citep{stylegan} from those generated by StyleGAN~\citep{stylegan}. This ease of separation exists not just for GANs but also for diffusion models~\citep{cozzolino2024raisingbaraigeneratedimage}. In short, no generative model can pass this test in the RGB space (yet). The common reason for this high accuracy is that fake images have certain isolated properties, e.g., presence of low-level checkerboard artifacts~\citep{zhang2019gan, cnn-detect} which the classifier utilizes to make its determination. However, these artifacts often remain imperceptible to the human eye. In fact, the overall quality of generated images has become visually so convincing that humans are struggling to identify what is real and what is fake~\citep{lu2023seeing, diel_human_judgement}. Could it be that these models are failing the test only because of low level artifacts, but for higher level properties – e.g., outline of a face, positioning of facial keypoints – these models are actually close to passing the two sample test? The question that we study, then, is how much we need to \textit{ignore} to reach that ultimate goal where even a neural network based classifier cannot distinguish real from fake. To study this, we create various levels of data abstractions where the original raw pixels of real and fake images are transformed into a representation space where some information gets lost. These operations include downsampling, cropping, blurring, converting into edge/semantic maps etc. For each of these abstractions, we train new classifiers to see if the transformed real and fake data are now separable. In doing so, we observe an unexpected phenomenon. In almost all cases, the classifier is able to obtain a high test accuracy (significantly above $50\%$). Fig.~\ref{fig:teaser} shows some examples of these data abstractions and corresponding accuracies. It turns out that the only means by which the classifier attains close to $50\%$ test accuracy is if we abstract away \emph{almost the entire information} localized in the original real and fake images; e.g., downsizing $256 \times 256$ resolution images to $4 \times 4$ (last column).

We conduct this study on a variety of real image datasets including MNIST~\citep{lecun-mnist1998}, CelebA~\citep{liu2015faceattributes}, FFHQ~\citep{stylegan} and GenImage~\citep{zhu2023genimage}, and fake images taken from various types of generative models like ACGAN~\citep{acgan}, DCGAN~\citep{radford-iclr16}, StyleGAN2~\citep{Karras2019stylegan2}, VD VAE~\citep{vdvae} and text-to-image diffusion models like Midjourney, Stable Diffusion~\citep{wukong, stable-diffusion, midjourney}. Across all these settings, the general trend holds in which the statistical indistinguishability, i.e., $\sim50\%$ test accuracy cannot be achieved until the representation level has abstracted away most of the information available in the input. We believe that our results should be surprising as they point out that the divide between real images and those produced by generative models is bigger than we typically imagine.

\section{Related work}\label{sec:related}

\paragraph{Generative model evaluation.} The field of generative models has been transformed in the last decade. From the original models which could barely generate handwritten digits \citep{goodfellow, vae}, we have come a long way where models can now generate almost anything, be it art, realistic renderings of natural scenes \citep{stable-diffusion, dall-e2, dalle-orig, ldm, midjourney}. Many metrics have been proposed which try to capture properties like image quality and diversity of the generated images. Inception score~\citet{salimans-nips16} and FID~\citet{fid} were one of the first popular metrics which encapsulate both of these aspects into one score. Other metrics were proposed which had the objective of disentangling these two aspects, to know how exactly a particular model is failing, or to have a better feature space than FID \citep{Kynkaanniemi2019, precision_recall_distributions, pmlr-v162-alaa22a, flaws_gen_metrics}. Another way to evaluate how close generated images are to real ones is through two sample test~\citep{two-sample}, where a classifier is trained to distinguish the two classes; if it cannot succeed at this task (accuracy remaining $\sim50\%$), then it can be inferred that the fake and real images come from exactly the same underlying distribution. For most generative models, this test in its original form always gives a very high accuracy (e.g., Fig.~\ref{fig:teaser} column 1). However, just because a generative model did not approximate the entirety of real image distribution perfectly does not necessarily imply that it cannot perfectly model \emph{some} aspect of that distribution. E.g., maybe the real and fake images are completely indistinguishable when considering only the edge details of the faces, or locations of important keypoints. In this work, we employ the classifier test to study if perfect statistical indistinguishability been achieved on some meaningful representations of images.

\paragraph{Properties of fake images and detection.} The image forensics community has tried to study if there are properties of fake images that can be used to distinguish them from real. Some have shown that the commonly used upsampling operations in a generative model often leave a high frequency trace in the generated image which can sometimes be seen with naked eyes as well \citep{zhang2019gan, odena2016deconvolution, ma_generalizable_artifacts}. Some works have shown where, spatially speaking, these artifacts reside most, e.g., in a person's hair~\citep{patchforensics}. Some works have used other higher level features like turning off high frequency component, representations from pretrained encoder, geometric properties of the scene~\citep{Sarkar_2024_CVPR, beyond_spectral, ojha2024universalfakeimagedetectors}. All of these different observations have then been used to deploy fake image detectors which can detect images not just from the model used during training, but also from unseen generative models \citep{cnn-detect, cozzolino_fake, ojha2024universalfakeimagedetectors, cozzolino2024raisingbaraigeneratedimage, rajan2025aligned, wang2023dire}. Our work is also concerned with studying the properties of fake images. However, the principal way in which our goal differs is that we wish to test if the errors in generative models are only due to certain isolated sources (e.g., checkerboard artifacts), or are they spread throughout in a much more holistic way. Furthermore, to make these detectors work in real world settings, past work has studied the sensitivity of trained detectors to different image manipulations and degradation operations like blurring, cropping, JPEG compression~\citep{cnn-detect, rajan2025aligned, GenD}. This has been aided in part by standardized datasets that test a detector's robustness to various post processing operations in a systematic way~\citep{DeepfakeBench_YAN_NEURIPS2023, li2025artificial}. The general conclusion from these works is that increasing the strength of image manipulation (e.g., blurring) hurts the detector's performance. While our work also deals with similar image manipulations, it serves a very different purpose. Our goal is not to build a single all-purpose classifier that can detect real/fake images from all representation level. Rather, we train \emph{new} classifiers from scratch for each representation level (different columns of Fig.~\ref{fig:teaser}) 
to see if the real and fake images becomes statistically indistinguishable at any such level.

\section{Approach}\label{sec:approach}
We will first briefly formalize what data is, using the example of MNIST. Let $\mathcal{R}$ denote the $N$ data samples that have been collected from the real world, e.g., by asking people to write digits on a paper, capturing the image and processing it. The statistical mind believes there to be a data generating process, and consequently views these recorded images as outcomes of an underlying probability distribution. Hence, if we denote $S$ as the bigger domain of all possible digits that can be written in reality, then $\mathcal{R} \sim S$. $G$ is trained on $\mathcal{R}$ to approximate the underlying distribution by mapping noise $z$, sampled from a known latent space, into a generated image. During test time, sampling a new, unseen $z$ and passing it through the trained $G$ to create a new image can conceptually be understood as collecting a new, unseen handwritten digit by a new person from the real world; a digit that was not originally present in $\mathcal{R}$. In other words, the generated images, represented as $\mathcal{F}$, should ideally be able to function as if the original real data collection process has been repeated, i.e., $\mathcal{F} \sim S$.

\subsection{Deep neural network classifiers for i.i.d detection}

To study this, we need a metric $d(\mathcal{R}, \mathcal{F})$ that gives a clear signal whenever $\mathcal{R}, \mathcal{F} \overset{\mathrm{iid}}\sim S$ regardless of the dataset type. FID, while generally helpful, is not the most appropriate metric in this case. To understand why, consider the dataset $\mathcal{R}$ being randomly shuffled and split in two halves $\mathcal{R}_1$ and $\mathcal{R}_2$. We know, by definition, that $\mathcal{R}_1, \mathcal{R}_2 \sim S$. If we set $\mathcal{R}$ as MNIST or FFHQ, FID($\mathcal{R}_1, \mathcal{R}_2$) will be different in magnitude. Hence there will be no way for us to confidently infer that $\mathcal{R}_1$ and $\mathcal{R}_2$ are actually iid samples in both cases. Hence, taking motivation from the line of work in image forensics community \citep{cnn-detect, ojha2024universalfakeimagedetectors}, and especially from \citet{two-sample}, we train various types of deep neural networks for binary classification to distinguish images from $\mathcal{R}$ and $\mathcal{F}$. If $\mathcal{R}, \mathcal{F} \sim S$, there should not be \emph{any} signal available for a classifier to use to separate them, at least on the test set. That is, the classifier's accuracy should only hover at $\sim50\%$ throughout the training process. Specifically, we split $\mathcal{R}$ and $\mathcal{F}$ into their respective training and test splits and train a classifier, $h$, on $\mathcal{D}_{train} = \{\mathcal{R}_{train} \cup \mathcal{F}_{train}\}$ and test it on $\mathcal{D}_{test} = \{\mathcal{R}_{test} \cup \mathcal{F}_{test}\}$. 

While $\mathcal{R}, \mathcal{F} \sim S$ necessitates $\sim50\%$ test accuracy of $h$ on $\mathcal{D}_{test}$, the inverse, i.e., chance performance on $\mathcal{D}_{test}$ does not necessarily imply $\mathcal{R}, \mathcal{F} \sim S$. There are two reasons for this. First, just because a single classifier, with its specific training configuration (e.g., network architecture, hyperparameters, weight initialization), cannot learn to separate $\mathcal{R}$ and $\mathcal{F}$ does not imply that a decision boundary does not exist. It is possible that a different $h$ could have found it. We provide a practical workaround (though, not a theoretical one) by training a family of neural net classifiers $\mathcal{H}$ and observing if any of them can get $>50\%$ test accuracy.

\begin{figure*}[t!]
    \centering
    \includegraphics[width=0.98\textwidth]{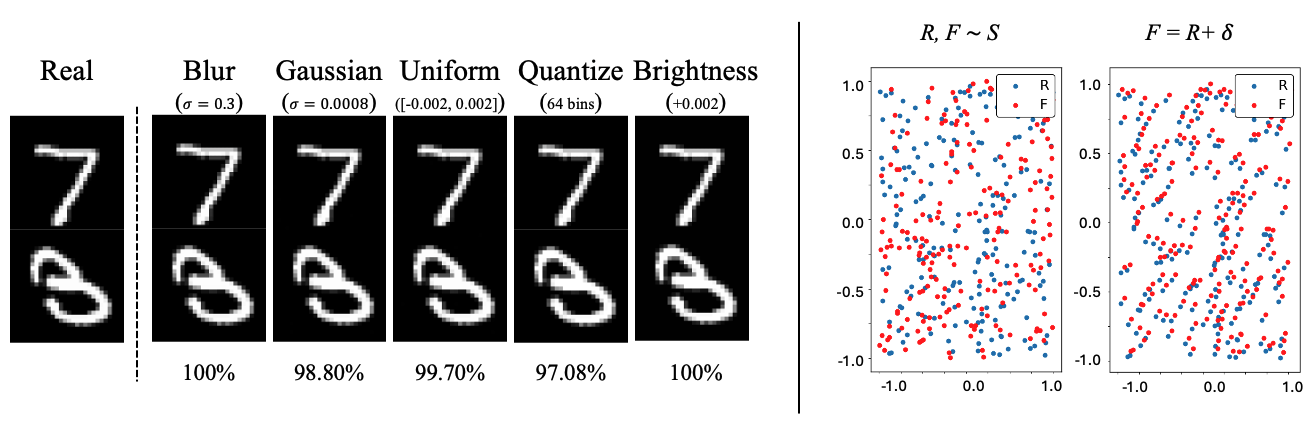}
    \vspace{-5pt}
    \caption{
\textbf{\textit{Left}}: even when visually \textit{mild} abstractions such as blur, Gaussian noise,
uniform noise, quantization, or brightness shifts are applied, binary neural
classifiers can still separate real and fake MNIST samples with high test
accuracy. \textbf{\textit{Right}}: toy examples contrasting the desired i.i.d. case,
$\mathcal{R},\mathcal{F}\sim S$, with a non-i.i.d. shifted fake distribution,
$\mathcal{F}=\mathcal{R}+\delta$, where the two sample sets are visually close
but statistically structured differently.
}
\label{fig:classifier_theory}
\vspace{-8pt}
\end{figure*}

Since neural networks have generally been used as classifiers when there is a visible semantic difference in the two classes of images (e.g., ImageNet, CIFAR), there might be some questions whether they can also be used in cases when only very subtle differences exist between the two classes. To study this, we design an experiment where we take MNIST as $\mathcal{R}$ and create different corrupted versions, $\mathcal{R}'$s, of it by (i) very slightly blurring it ($\sigma=0.3$), (ii) adding small Gaussian noise ($\sigma = 0.0008$), (iii) quantizing the images into 64 bins (original had 256), (iv) altering just a few pixels with small salt and pepper noise. The resulting images are shown in Fig.~\ref{fig:classifier_theory} left, where we see that there is barely any difference between real and corrupted versions of it.
For each case, we train ResNet~\citep{resnet} on each $\mathcal{D}_{train} = \{\mathcal{R}_{train} \cup \mathcal{R}'_{train}\}$ and test on the corresponding $\mathcal{D}_{test}$. The results are shown at the bottom of Fig.~\ref{fig:classifier_theory}, where see that the classifier consistently gets $>90\%$ test accuracy. This stress test gives us some confidence that if there is a pattern to distinguish two classes of images, regardless of how subtle it is, at least some $h \in \mathcal{H}$ will be able to find it. More formally, the metric that we were looking for can be defined as following:

\begin{equation}
    d_{\mathcal{H}}(\mathcal{R}, \mathcal{F}) = \max_{h \in \mathcal{H}}|\mathbb{E}_{\mathcal{R}}[h(X)] - \mathbb{E}_{\mathcal{F}}[h(X)] |
    \label{eq:metric_clean}
\end{equation}

$d_{\mathcal{H}}(\mathcal{R}, \mathcal{F})$ will be zero if and only if none of the classifiers $h \in \mathcal{H}$ can learn to separate $\mathcal{R}$ and $\mathcal{F}$. 

However, even if $\mathcal{H}$ could contain all possible classification
functions, $d_{\mathcal{H}}(\mathcal{R}, \mathcal{F}) \sim 0$ still cannot
guarantee $\mathcal{R}, \mathcal{F} \sim S$. Figure~\ref{fig:classifier_theory}
(right) explains why. Case (a) depicts an when all the samples across $\mathcal{R}$ and $\mathcal{F}$ are i.i.d, whereas in case (b) $\mathcal{F} = \mathcal{R} + \delta$ where $\delta \sim 0$, i.e., not independent. In both cases, no function $h$ can perform well on the respective $\mathcal{D}_{test}$. The question of quantitatively differentiating case (a) from (b) is an open one, and we leave it as such. Fortunately, as we will see in the results, we do not have to deal with this scenario for the most part. If the metric defined in Eq.~\ref{eq:metric_clean} is positive, we can be sure that $\mathcal{R}$ and $\mathcal{F}$ are not iid samples from $S$.

\begin{figure*}[t!]
    \centering
    \includegraphics[width=1.0\textwidth]{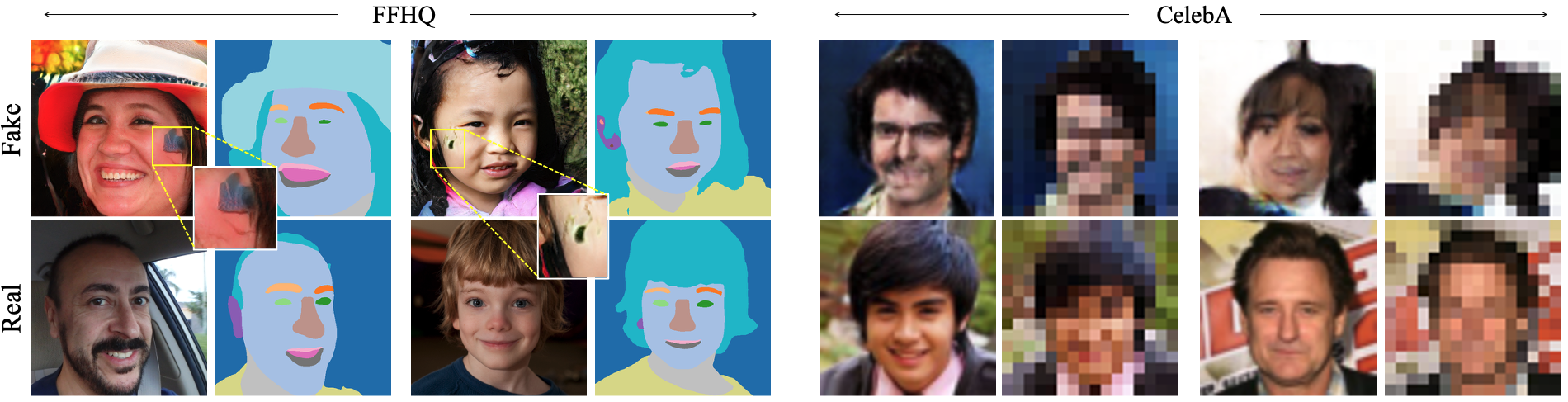}
    \caption{
Motivation for image abstractions. Left: fake FFHQ images may contain localized
artifacts, such as unnatural skin, that disappear after
mapping the image to a higher-level semantic segmentation representation.
Right: CelebA examples show a spatial abstraction, where downsampling removes
fine image details while preserving coarse facial structure. We apply such
abstractions to both real and fake images and test whether classifiers can still
distinguish $f(\mathcal{R})$ from $f(\mathcal{F})$.
}
\label{fig:abstract_motivation}

\end{figure*}

\subsection{Search for i.i.d through data abstractions}
We know from prior work that generative models can introduce errors for various reasons \citep{zhang2019gan, odena2016deconvolution}. A common example is checkerboard artifact due to the upsampling process. Beyond these, the images can often contain some semantic anomalies (e.g., weird lips, presence of six fingers) \citep{patchforensics}. But at the same time those images, notwithstanding the errors, appear good enough in other aspects, and also novel. 
\begin{center}
\textit{Could it be the case that $\mathcal{R}, \mathcal{F} \overset{\mathrm{iid}}\sim S$ if some small errors in $\mathcal{F}$ could be ignored?} 
\end{center}

To study this, we propose to create abstracted versions of the original RGB real/fake images so that some information gets lost, with the hope of also washing away the errors by the generative model. Figure~\ref{fig:abstract_motivation} explains the intuition. Taking the example of FFHQ, we notice that the original fake image has a weird artifact on the face, something that is not expected to be found in a real image. As an example abstraction, we extract segmentation masks of both real and fake images. In this space, that artifact is no more as that information is abstracted away. There are many other ways to do this as well. In the context of images, it is possible to preserve the resolution and apply operations such as blurring, adding noise, quantization, converting to edge/semantic maps etc. We can also reduce the dimensionality by operations such as downsizing, cropping. All of these provide avenues to hide the errors of $G$.

Formally, let $f$ denote the function which transforms an image $X$ into its abstracted version $f(X)$. After each such function application on both $\mathcal{R}$ and $\mathcal{F}$, we train a new collection of classifiers $\mathcal{H}$ from scratch. Extending the definition in Eq.~\ref{eq:metric_clean}, our generalizable metric becomes:

\begin{equation}
    d_{\mathcal{H}, f}(\mathcal{R}, \mathcal{F}) = \max_{h \in \mathcal{H}}|\mathbb{E}_{\mathcal{R}}[h(f(X))] - \mathbb{E}_{\mathcal{F}}[h(f(X))] |
    \label{eq:metric_clean_v2}
\end{equation}

In our experiments, we will search over the space of $f$ to see if $d_{\mathcal{H}, f}$ ever becomes $\sim0$.

\section{Experiments}

We now explain the specific experiments we conduct to test if/when $\mathcal{R}, \mathcal{F} \overset{\mathrm{iid}}\sim S$. In total, we consider four datasets of real images of increasing complexity \& resolution; MNIST~\cite{lecun-mnist1998} ($28 \times 28$), CelebA~\cite{liu2015faceattributes} ($64 \times 64 \times 3$), FFHQ~\cite{stylegan} (we consider $256 \times 256 \times 3$ version) and GenImage~\cite{zhu2023genimage} ($256 \times 256 \times 3$). For each of these datasets, we gather corresponding fake images by either training appropriate generative models ourselves or by sampling from official pre-trained models.

\begin{figure*}[t!]
    \centering
    \includegraphics[width=1.0\textwidth]{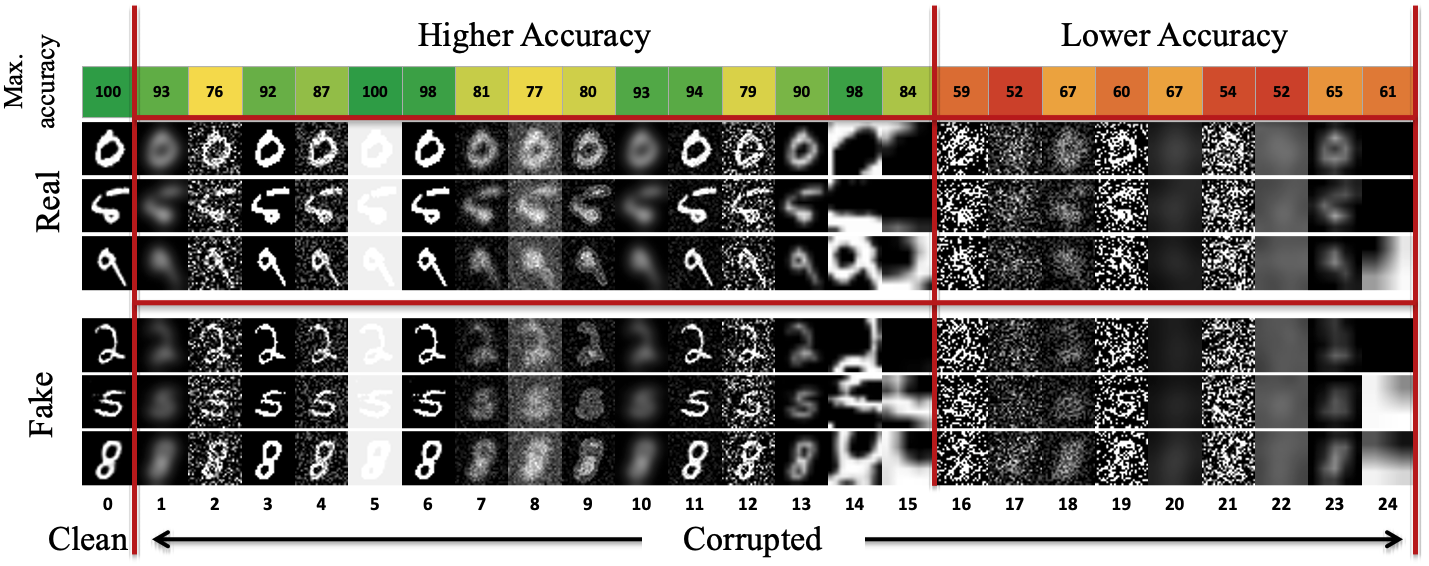}
    \caption{
MNIST real-vs-fake detection under image abstractions. Columns show clean and corrupted versions of matched real and ACGAN-generated digits, while the heatmap reports the best test accuracy across DenseNet-Small, Mini-Xception, and ResNet-Small. Real and fake images remain separable through many abstractions and approach chance only when the digit content is heavily destroyed. Please see Table~\ref{tab:mnist_25_configs} for corruption configurations.
}
\label{fig:mnist_results}

\end{figure*}

\subsection{MNIST}\label{sec:mnist}
The real dataset, $\mathcal{R}$, is categorical in nature, having a total of $70$k images distributed among the training ($\mathcal{R}_{train}$), validation and testing $\mathcal{R}_{test}$ splits of $50$k, $10$k and $10$k images respectively. Each class ($0-9$) has the same number of images. We train ACGAN~\cite{acgan} on $\mathcal{R}_{train}$, whose latent space consists of two inputs, a continuous noise vector $z \sim \mathcal{N} (0,1)$ and a categorical code $c \sim \text{Cat}(K = 10,p = 1/10)$. We pass these inputs the generator to obtain a fake image $G(z,c)$. For each $c$ (representing a digit category), we generate $5000$ images by sampling that many $z$. This makes sure that $|\mathcal{R}_{train}| = |\mathcal{F}_{train}|$. Similarly, we generate $\mathcal{F}_{test}$ ensuring $|\mathcal{R}_{test}| = |\mathcal{F}_{test}|$. Our $\mathcal{H}$ consists of three models; ResNet, DesnseNet~\cite{DenseNet2017} and mini-Xception~\cite{Chollet2016XceptionDL}. We train all these classifiers $h$ on $\mathcal{D}_{train} = \{\mathcal{R}_{train} \cup \mathcal{F}_{train}\}$ for 25 epochs, and test it on $\mathcal{D}_{test} = \{\mathcal{R}_{test} \cup \mathcal{F}_{test}\}$. This process is repeated separately for each abstraction level. 

We consider the following abstraction operations $f$: (i) identity function (original real/fake images), (ii) blurring (with different $\sigma$), (iii) adding Gaussian/uniform/salt and pepper noise, (iv) quantization, (v) downsampling, (vi) cropping. In Fig.~\ref{fig:mnist_results}, we show many combinations of these, and the details of these configurations are present in Table~\ref{tab:mnist_25_configs}. We also show the test accuracy across ResNet, DesnseNet and mini-Xception. 

We begin the observation from left, i.e., original images, and see that all classifiers can easily separate them, which indicates the presence of at least some errors. This echoes the general finding of image forensics community. However, some errors seem to persist as we continue to apply different types and levels of abstraction. For example, even after blurring $\mathcal{R}$ and $\mathcal{F}$ with $\sigma=5.0$, one of the $h \in \mathcal{H}$ is still able to separate them $80\%$ of the time. Another example is downsizing the image to a $5 \times 5$ matrix, and a classifier can still identify ${2/3}^{rd}$ of the images. 
%As a final example, we see that even if we reduce the original dimension by \emph{$\sim 100$ times} using PCA ($784 \rightarrow 8$), a classifier can still detect $76\%$ of the images. 
Note that all of these operations are abstracting away information in different ways. This means that the nature of error left in $\mathcal{F}$ (e.g., blurring vs downsizing), which gets picked up by $h$, is different.  

On the right side of Fig.~\ref{fig:mnist_results}, we show those cases of $f$ where all the classifiers perform only at around chance performance ($50-55\%$). The key thing to notice is the effect these $f$'s have on the original image. The semantic content seems almost completely lost to the human eye. In other words, to have $\mathcal{F}$ become statistically indistinguishable from $\mathcal{R}$, one needs to, e.g., blur the image to such an extent, or make its resolution so small, or add so much noise, that the original digit in either the real or fake image is barely visible. We will discuss the implications of this later on in the paper.

\begin{figure*}[t!]
    \centering
    \includegraphics[width=1.0\textwidth]{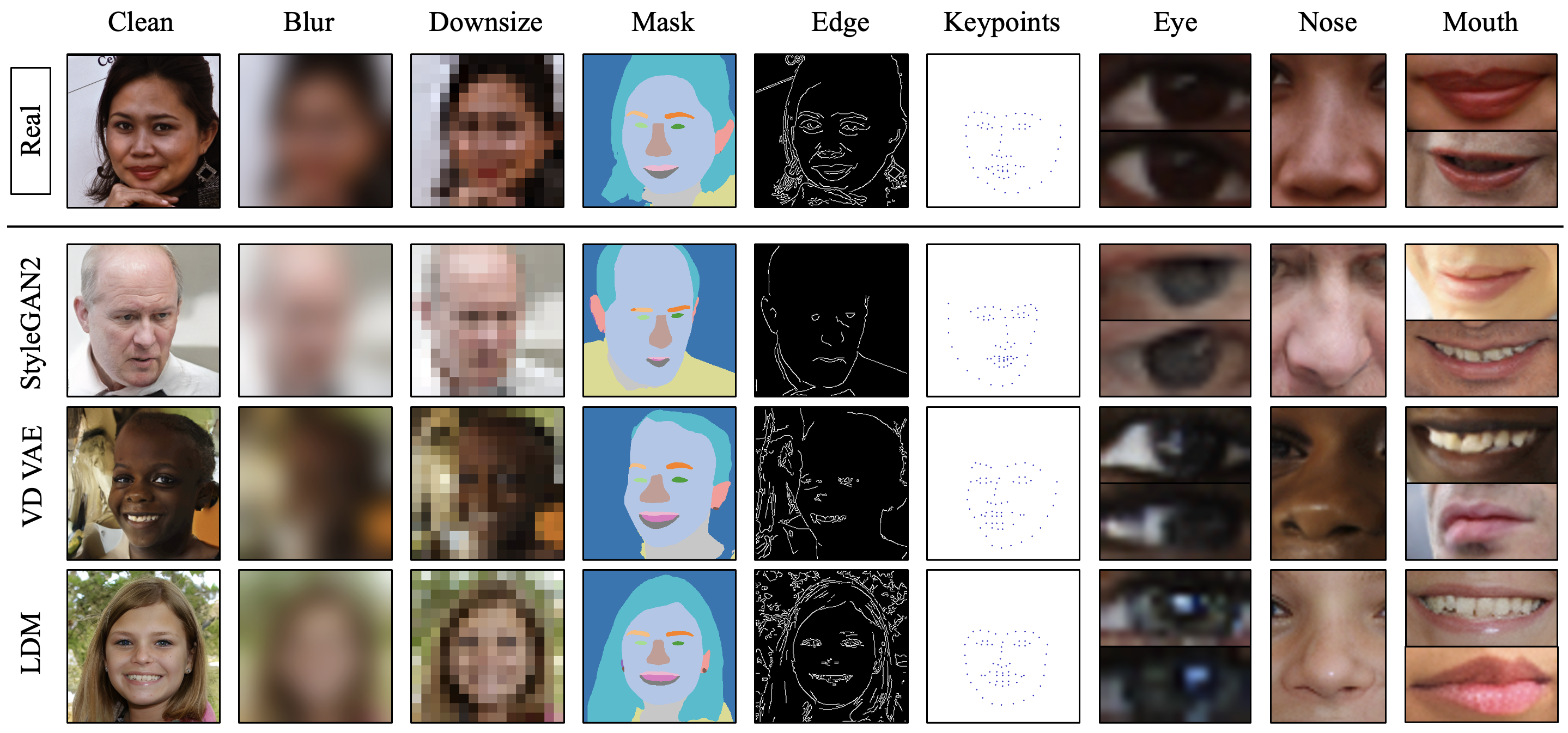}
    \caption{FFHQ real (top) and StyleGAN2, VD VAE, LDM-generated faces under various data abstractions. Even after abstracting away enough information from real and fake images, e.g., by converting them into semantic mask, we do \emph{not} observe statistical indistinguishability, as a classifiers trained from scratch are able to obtain accuracies way above $50\%$ in Table~\ref{tab:ffhq_main}.  
    }
    \label{fig:ffhq_results_all}
\end{figure*}

\subsection{FFHQ}\label{sec:ffhq}
Next, we experiment with the dataset whose release, along with the associated generative model, led to a genuine confusion among humans in being able to recognize what is real or fake. FFHQ comprises of $70$k high quality images of human faces. We consider the $256 \times 256$ resolution version of the dataset and experiment with three different types of generative models trained on this dataset – StyleGAN2~\cite{Karras2019stylegan2}, VD VAE~\cite{vdvae} and latent diffusion models (LDM)~\cite{ldm}. From each of the generative model, we generate $70$k fake images at the same resolution without any truncation, thus making $|\mathcal{R}| = |\mathcal{F}|$. This dataset is split into $80/20$ train/test split. We use ResNet, DenseNet and ViT as $\mathcal{H}$ and train them separately for (i) FFHQ vs StyleGAN2, (ii) FFHQ vs VD VAE, and (iii) FFHQ vs LDM.

We begin by considering three types of abstractions which can vary in strength; (i) blurring, (ii) downsizing and (iii) cropping. Fig.~\ref{fig:ffhq_results_all} (columns 1-3) shows the effect, and Table ~\ref{tab:ffhq_main} shows the results. \emph{To the human eye}, blurring ($\sigma=20$) both real and fake images, or downsizing them to $16 \times 16$ resolution might appear to have removed all differences between the two. However, the classifier's performance remains way above $\sim50\%$ for all the generative models. In other words, some unexpected signal still remains that the classifier uses to separate the two classes. In fact, to arrive at a state where the classifier cannot find much difference ($50-55\%$), we need to abstract to such an extent that the contents of the original image are virtually completely gone; e.g., downsizing/cropping to $1 \times 1$ region. Please see Fig.~\ref{fig:ffhq_results_stylegan} in the appendix for a more detailed qualitative visualization of these abstraction processes. 

%There is a point worth noting though. There are multiple ways to arrive at a $4 \times 4$ compressed version of a $256 \times 256$ image; we can downsize the whole image, or we can extract the central $4 \times 4$ pixels. Both result in different amounts of fake signal for a classifier to use (maximum test accuracies are $96.3\%$ and $58.4\%$ respectively). So, even though the downsized version provided more information about the whole image, the more relevant factor for fakeness, in this case, is the availability of original raw pixels. 

\begin{table*}[]
\resizebox{\textwidth}{!}{
\begin{tabular}{lcccccccccccc}
\toprule
\multirow{2}{*}{\textbf{Model}}
& \multicolumn{4}{c}{Downsizing}
& \multicolumn{4}{c}{Cropping}
& \multicolumn{4}{c}{Blurring}
\\
\cmidrule(lr){2-5} \cmidrule(lr){6-9} \cmidrule(lr){10-13} 
& $64{\times}64$ & $16{\times}16$ & $4{\times}4$ & $1{\times}1$
& $64{\times}64$ & $16{\times}16$ & $4{\times}4$ & $1{\times}1$
& $\sigma{=}5$ & $\sigma{=}10$ & $\sigma{=}20$ & $\sigma{=}40$ \\
\midrule
StyleGAN2
& 99.8\% & 86.2\% & 58.4\% & 52.6\%
& 100\% & 99.8\% & 96.3\% & 54.8\%
& 98.0\% & 93.0\% & 74.0\% & 63.0\% \\
VD VAE
& 98.9\% & 91.9\% & 69.3\% & 57.0\%
& 100\% & 99.8\% & 85.9\% & 57.8\% 
& 95.8\% & 93.2\% & 83.0\% & 73.6\% \\
LDM
& 99.5\% & 73.7\% & 58.1\% & 54.7\%
& 100\% & 99.6\% & 82.5\% & 55.5\%
& 93.0\% & 79.6\% & 64.8\% & 59.1\% \\
\bottomrule
\end{tabular}
}
\caption{FFHQ real-vs-fake test accuracy of the best performing classifier under various data abstractions for StyleGAN2, VD VAE and LDM. Detection remains high for many spatially degraded inputs and approaches chance only under the strongest abstractions, such as $1\times1$ spatial collapse.}
\label{tab:ffhq_main}
\end{table*}

\begin{table}[h]
\centering
\resizebox{0.85\textwidth}{!}{%
\begin{tabular}{lccccccccc}
\toprule
& Edge    &  Mask    &  Keypoint   &  Eye    &  Nose    &  Mouth    &  L-eye    &  R-brow    &  R-ear        \\
\midrule
StyleGAN2   & 91.7\% & 76.1\% & 78.3\% & 99.8\% & 99.9\% & 99.9\% & 99.2\% & 98.9\% & 99.0\% \\
VD VAE     & 95.0\% & 70.3\% & 76.5\%  & 99.9\% & 99.9\% & 99.9\% & 99.9\% & 99.9\% & 99.9\% \\
LDM        & 99.8\% & 67.8\%  &  75.0\%  & 99.9\% & 100.0\% & 100.0\% & 99.9\% & 99.9\% & 99.9\% \\
\bottomrule
\end{tabular}
}
\caption{ FFHQ real-vs-fake test accuracy of best performing classifier for StyleGAN2, VD VAE and LDM across various data abstractions (first 3 columns) and cropped face parts (last 6 columns). }
\label{tab:ffhq_v2}
\end{table}

Beyond these operations, we experiment with other abstractions which are semantically more meaningful given the high resolution nature of images. First, we use Canny edge detection to obtain the edge boundaries of real \& fake images, which masks away, among other things, color information. Second, we convert the RGB image into its semantic mask using a pretrained segmentation model, something which conceptually is going to hide the high frequency noise, and potentially also hides semantic anomalies as explained in Fig.~\ref{fig:abstract_motivation}. Next, we use a pretrained keypoint detector to extract 68 landmarks from a real/fake face image. This operation goes beyond the previous two in masking any explicit outline/edge in an image. The output from these operations (semantic mask, edge map, 68 landmark locations) gets fed into a classifier which is then trained to separate the real and fake class. 
%The output is single channel image which we normalize and feed into the classifiers described above. We also experiment with converting the RGB image into its semantic mask use a pretrained segmentation model, something which conceptually is going to hide the high frequency noise. The output is a single channel matrix of $256 \times 256$ resolution having class prediction index from $[0-18]$ (e.g., skin, nose, eye) for each pixel location. We normalize this and feed it into the same classifiers. 
Lastly, using the same semantic masks, we crop out certain regions of the face, e.g., its eyes, mouth, nose, ear, for both real and fake images. This way, we can use the classifiers in $\mathcal{H}$ to distinguish between real and fake eyes, or real and fake mouth etc. More details about the data format for semantic mask, edge map, landmark locations and details about the cropping mechanism is explained in the appendix Sec. \ref{app:ffhq}.

Figure~\ref{fig:ffhq_results_all} (columns 4-9) shows examples of the semantic and local-part abstractions, and Table~\ref{tab:ffhq_v2} summarizes the classification results. In summary, none of these operations are enough to bring the test accuracy of all the classifiers to $\sim50\%$. The implication is that the edges of fake images are different than real ones, fake eyes are different from real ones, the locations of various keypoints of fake images are different from real images etc. Similar to Table~\ref{tab:ffhq_main}, this trend holds more or less similarly for all the three models. This principle of needing to abstract away almost the entire information of the original image to achieve statistical indistinguishability is observed also for another dataset of human faces – CelebA~\cite{liu2015faceattributes}, for which we present the results in appendix Sec.~\ref{app:celeba}.

\subsection{GenImage}\label{sec:genimage}

\begin{figure*}[t!]
    \centering
    \includegraphics[width=1.0\textwidth]{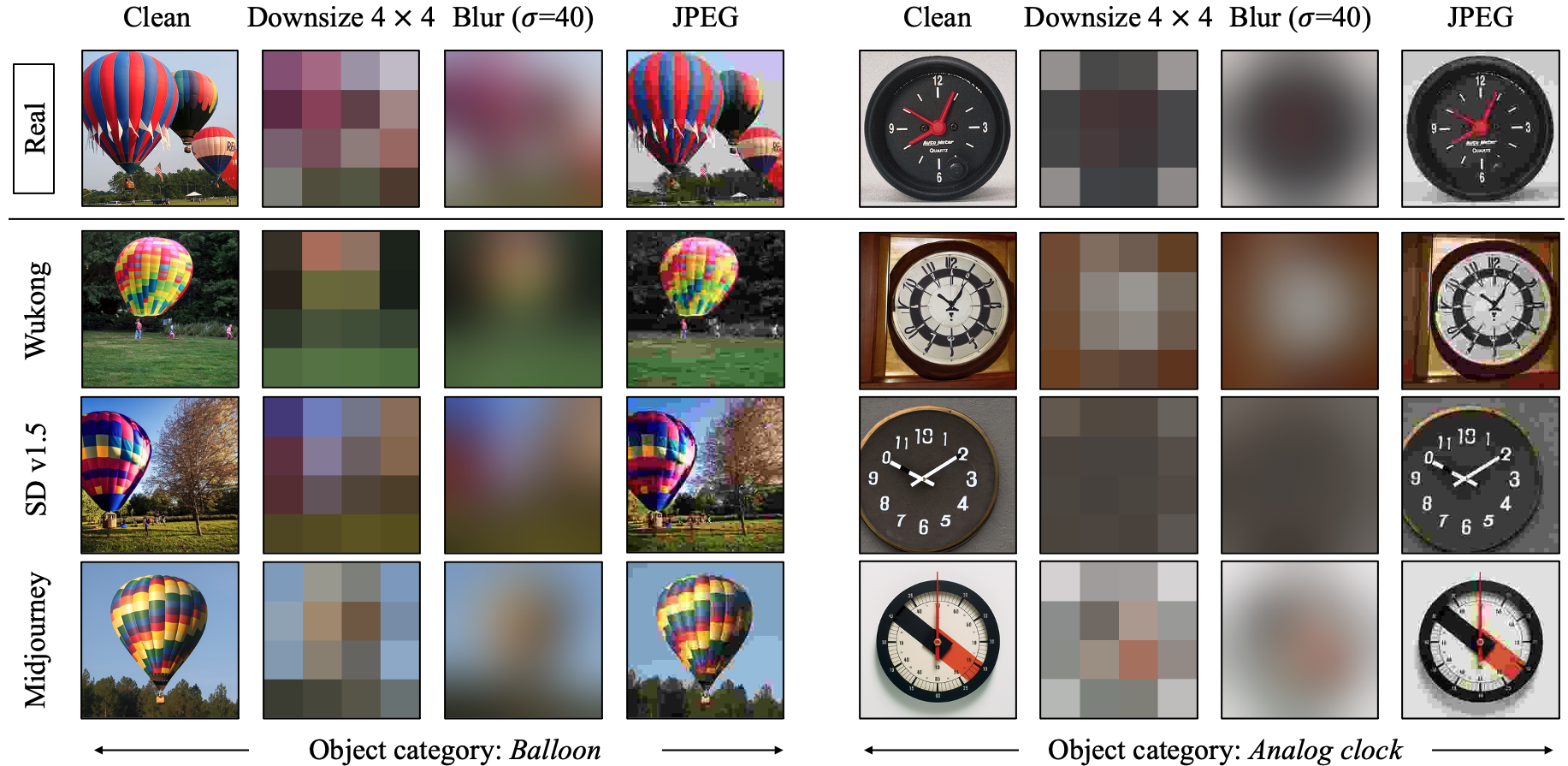}
    \caption{Real (top) and fake images (bottom three) from GenImage dataset under different spatial and blur abstractions. The original images are of $256 \times 256$ resolution. Even after abstracting the information content to a great extent by downsizing, blurring or compressing, the resulting real and fake images are still distinguishable by a classifier trained from scratch, as shown in Table~\ref{tab:genimage_main}.}
    \label{fig:genimage_results}
\end{figure*}

\begin{table*}[t]
\resizebox{\textwidth}{!}{
\begin{tabular}{lcccccccccccc}
\toprule
\multirow{2}{*}{\textbf{Model}} 
& \multicolumn{3}{c}{Downsizing} 
& \multicolumn{3}{c}{Cropping} 
& \multicolumn{3}{c}{Blurring} 
& \multicolumn{3}{c}{JPEG} 

\\
\cmidrule(lr){2-4} \cmidrule(lr){5-7} \cmidrule(lr){8-10} \cmidrule(lr){11-13}  
& $16 \times 16$ & $4 \times 4$ & $1 \times 1$ & $16 \times 16$ & $4 \times 4$ & $1 \times 1$ & $\sigma=10$ & $\sigma=20$ & $\sigma=40$ & 50 & 25 & 5 \\
\midrule
Wukong 
& 90.9\% & 81.1\% & 66.7\%
& 97.5\% & 66.2\% & 61.9\%
& 91.4\% & 87.2\% & 83.9\%
& 96.6\% & 96.0\% & 92.6\%
 \\
SD v1.4
& 91.1\% & 80.9\% & 68.6\%
& 76.1\% & 61.5\% & 58.7\%
& 91.1\% & 87.5\% & 82.4\%
& 97.5\% & 94.8\% & 75.9\%
\\
SD v1.5
& 89.7\% & 80.2\% & 67.8\%
& 75.6\% & 61.2\% & 58.4\%
& 89.9\% & 86.0\% & 81.2\%
& 96.9\% & 95.8\% & 91.1\%
\\
Midjourney 
& 78.4\% & 69.1\% & 58.7\%
& 93.2\% & 73.5\% & 57.4\%
& 81.2\% & 74.7\% & 70.3\%
& 92.7\% & 90.2\% & 84.3\%  

\\

\bottomrule
\end{tabular}
}
\caption{GenImage real-vs-fake test accuracy under downsampling, cropping, blurring and JPEG abstractions. Detection remains well above chance for
most abstractions and approaches chance only under extreme spatial collapse,
such as $1\times1$ downsampling or cropping.}
\label{tab:genimage_main}
\end{table*}

Finally, we experiment with more real world settings going beyond single object categories of handwritten digits or human faces. GenImage~\cite{zhu2023genimage} was proposed for the task of fake image detection, and consists of real images from ImageNet~\cite{imagenet} and fake images generated by popular text-to-image models. For our experiments, we consider images from four models; Wukong~\cite{wukong}, Stable Diffusion (SD) v1.4 and v.15~\cite{stable-diffusion} and Midjourney~\cite{midjourney}. The real image dataset has $\sim168,000$ images roughly evenly distributed across the 1000 object categories of ImageNet. Each of the fake image datasets also has the same number of images across the same  categories. We use DenseNet as our classifier and train it separately for each generative model for binary classification, i.e., (i) ImageNet vs Wukong, (ii) ImageNet vs SD v1.4, (iii) ImageNet vs SD v1.5, and (iv) ImageNet vs Midjourney. The resolution of images is $256 \times 256$. For data abstraction, we borrow the operations described in Sec.~\ref{sec:ffhq} who strength can be varied – Downsizing, Cropping, Blurring. We add JPEG compression to this since it gets used in real world often. More details on GenImage dataset in appendix Sec. ~\ref{app:genimage}.

Fig.~\ref{fig:genimage_results} shows sample real and fake images from two object categories and the effect of different data abstraction operations (samples from more categories are shown in Fig.~\ref{fig:genimage_results_v2}). The results of training the classifiers on these settings is shown in Table~\ref{tab:genimage_main}. Similar to the trend that we have seen so far, the test accuracy remains consistently high across many cases. The reason this should be surprising is because according to common understand, the errors or artifacts of the generative model manifest primarily in the high frequency components of a generated image~\cite{zhang2019gan, cnn-detect}. If so, an operation like downsizing to $4 \times 4$ or extremely high blurring ($\sigma = 40$) or very high JPEG compression (columns 2, 3, 4 respectively) should eliminate the artifact, which should result in statistical indistinguishability, i.e., test accuracy of $\sim50\%$. However, our results suggest that each time the classifier can pick up some new signal to distinguish the two classes. Among the four models studied, Midjourney is the best (lower overall test accuracy). Collectively, these results across all of the datasets point to an understanding of fake images that we believe has been missing so far. We discuss this in the next section.

%%%%%%%%%%%%%%%%%%% BEGIN : ImageNet + complex models result %%%%%%%%%%%%%%%%%%%%%%%%%%%%%

%%%%%%%%%%%%%%%%%%% END : ImageNet + complex models result %%%%%%%%%%%%%%%%%%%%%%%%%%%%%

\subsection{Analysis}\label{sec:analysis}
We will focus on faces (FFHQ) since that has been the scope of much discussion in the community, but the general conclusions, we believe, apply to other domains such as MNIST and GenImage as well. What exactly is it that makes an image fake? Different works in the past have offered different perspectives. A line of work has attributed the fakeness to artifacts introduced due to the upsampling process. Within the context of images, the authors in \cite{patchforensics} show that \textit{hair, weird background} of the face is the source for fakeness. 

However, while these can certainly be legitimate sources of artifacts, our results seem to suggest that the nature of artifacts extends much beyond these isolated factors. The errors of $G$ cannot be said to reside only in the high frequency checkerboard artifacts; because $\mathcal{R}$ and $\mathcal{F}$ differ in their semantic maps too. The errors also cannot reside only in the inability to model color details properly, because $\mathcal{R}$ and $\mathcal{F}$ differ in their (colorless) edge maps too. For faces, the errors can also not be contained just in the hair of the subject or the distorted background, because the eyes/mouth/nose of $\mathcal{R}$ and $\mathcal{F}$ differ too. In fact, the errors cannot just be some semantic anomaly in some part, because $\mathcal{R}$ and $\mathcal{F}$ differ even if we blur or downsize (e.g., $16 \times 16$) them exceedingly. The general conclusion from our experiments is that it is very difficult to find a \emph{meaningful} aspect of the fake image within which it can be considered identical with real counterpart; i.e., $\mathcal{F}, \mathcal{R} \sim S$. The word `meaningful' is important because them being statistically indistinguishable only when downsized to a single pixel, or after adding so much noise that the original signal seems lost to the human eye -- is not very helpful.

% Although one can argue that the quality of dcgan is not that great, our main point is to focus on these cases where human fooling rate is nonetheless reached. What about these things. 

% , one cannot say it is only in structure of the face (downsampled images are different too), one cannot say that only hair is fake (eyes are fake too). the point being, it is difficult to find a property of the fake image which is iid with real. So, if error is to be defined as the "trace" of the generative model, this trace encomapsses almost every concievable property of the image. 
\section{Limitations}\label{sec:limitations}
Our work primarily relies on classifier test to study whether perfection can be achieved by generative models. However, as explained in Fig.~\ref{fig:classifier_theory}, while an accuracy of $\sim50\%$ by multiple classifiers can hint at statistical indistinguishability, it still does not prove that $\mathcal{R}, \mathcal{F} \sim S$. Hence, right now we do not properly understand those cases where $\mathcal{R}$ and $\mathcal{F}$ do indeed become indistinguishable. Another limitation concerns the potential usage for our method to rank different generative models. For example, model A can be considered better than model B if A achieves statistical indistinguishability for a data abstraction sooner than B. However, computing such ranking will be much more computationally heavy than other available metrics like FID. Finally, our work has shown results on single-object datasets but not for cases where both real and fake images depict more complex scenes (e.g., traffic).

\subsection*{AI use statement}
In this work, we used generative AI tools for generating synthetic datasets, implementing methods, supporting qualitative and thematic data analysis. We have not used generative AI tools for designing or providing feedback on research  methodology or experiments, interpreting results. Helping develop theoretical models or conceptual frameworks, formulating mathematical claims, providing critical ingredients for proving mathematical claims, assisting in the writing of proofs, proposing or refine hypotheses, assisting with translation, clean and reformat dataset are not applicable to this work. Additionally, we used generative AI tools for creating or editing software code, draft parts of the appendix section (the main paper has no AI usage). We have reviewed all AI-assisted work. The main parts of the code generated by AI has been verified by human authors for their accuracy. Some parts of the AI-generated code has been verified by multiple authors. We take responsibility for the final content of this work, including text, claims or artifacts produced with the aid of generative AI.

\subsection*{Ethics statement}
We believe that our work puts forward a good news for the image forensics community because it calms the apparent threat by generative models. As we have explained throughout the paper, if it was true that generative models, apart from some small mistakes here and there, were getting close to perfection, then there might have been a genuine cause for worry because after those small errors get fixed, there will be nothing one could have done to verify whether some digital content is real or AI-generated. However, our work hints that these models are much further away from the natural data than what was previously imagined.

\subsection*{Reproducibility statement}
In this work, the learning algorithm are a collection of binary classifiers. The main ingredient is what the data is on which these classifiers are trained on. To that end, we have included the details of how the real and fake images are gathered for each datasets in the beginning of Sec.~\ref{sec:mnist}, \ref{sec:ffhq} and \ref{sec:genimage}, including their number, sampling process etc. The details of the data abstraction are given in Sec.~\ref{app:mnist_abstractions}, \ref{app:celeba}, \ref{app:ffhq}, ~\ref{app:genimage}. The details about the classifier are given in Table~\ref{tab:mnist_classifier_architectures}, \ref{tab:celeba_classifiers} and \ref{tab:ffhq_classifiers}. We will open source the code upon publication.

%\subsubsection*{Acknowledgments}
%Use unnumbered third level headings for the acknowledgments. All acknowledgments, including those to funding agencies, go at the end of the paper.

\bibliography{iclr2027_conference}
\bibliographystyle{iclr2027_conference}

\appendix
\section*{Appendix}

\section{Compute Resources}
\label{sec:compute}

All experiments were run on a high-performance computing cluster using the SLURM workload manager. Training jobs were submitted primarily to the \texttt{Quick} GPU partition. Each training run used a single scheduled GPU, 4 CPU cores, and 16\,GB of host RAM unless otherwise noted. The available GPU pool includes NVIDIA A100, A40, Titan RTX, Titan X, Titan V, and GTX 1080 Ti nodes; the exact GPU model was assigned by the cluster scheduler and was not pinned per experiment. All reported runs fit within the memory of a single GPU. Datasets, checkpoints, logs, and generated images were stored on the cluster data filesystem.

\begin{table}[h]
\centering
\small
\begin{tabular}{l p{0.40\linewidth} c}
\toprule
Resource & Description & Usage \\
\midrule
Compute environment &
Institutional HPC cluster managed with SLURM.&
On-premise HPC \\

Accelerators &
NVIDIA GPU nodes scheduled from the pool, including A100, A40, Titan RTX, Titan X, Titan V, and GTX 1080 Ti GPUs. &
1 GPU per run \\

CPU and memory &
Standard training allocation used by the reported experiments. &
4 CPU cores, 16\,GB RAM \\

Storage &
Cluster data filesystem used for datasets, checkpoints, generated samples, logs, and analysis outputs. & personal env \\

Software environment &
Conda environment with PyTorch and torchvision; jobs launched through SLURM scripts. &
Python/PyTorch \\
\bottomrule
\end{tabular}
\caption{Compute environment used for the experiments.}
\label{tab:compute_environment}
\end{table}

Individual training runs were lightweight: MNIST classifier runs completed within the short \texttt{Quick} partition allocation using a single GPU, while the larger image-generation and face-dataset experiments were also run as independent single-GPU jobs. The total compute for the experiments reported in the paper is estimated to be on the order of tens of GPU-hours. The full research project consumed more compute than the final reported experiments because it included preliminary controls, architecture searches, corruption sweeps, failed or superseded runs, and pilot GAN experiments that were used to refine the final experimental design.

\section{Training details}
\label{sec:training-details}

All experiments involving CelebA and FFHQ dataset use binary cross-entropy loss (BCELoss), an 80/20 train/test split,
random seed 42, and a ReduceLROnPlateau learning rate scheduler
(factor $= 0.5$, patience $= 5$).
Labels are real $= 0$, generated $= 1$.
All models output $P(\text{real}) \in [0,1]$ via a sigmoid activation.
Images are normalized to $[-1, 1]$ via $(x_{\text{uint8}} / 255 - 0.5) / 0.5$.

%======================================================
\subsection{MNIST Experimental Details}
\label{app:mnist_details}

\subsubsection{ACGAN Training and Fake Dataset Construction}
\label{app:mnist_acgan}

The MNIST generator is an auxiliary-classifier GAN
(ACGAN) conditioned on the digit class.
The generator input is a $100$-dimensional latent vector
$z \sim \mathcal{N}(0,I)$ and a categorical digit label
$c \in \{0,\ldots,9\}$. The class label is embedded into a
$100$-dimensional vector and multiplied element-wise with $z$ before
being passed through the generator.

The generator maps the conditioned latent code through a fully-connected
layer followed by two upsampling convolutional blocks and a final
\texttt{tanh} layer producing $28{\times}28$ grayscale images. The
discriminator is a four-stride-2 convolutional stack with two heads:
an adversarial real/fake head trained with binary cross-entropy and an
auxiliary digit-class head trained with cross-entropy. Both networks are
optimized with Adam (learning rate $2\times 10^{-4}$, $\beta_{1}=0.5$,
$\beta_{2}=0.999$), batch size $64$, until the visual quality of
generated images saturates (\(\sim 200\) epochs in practice).

For downstream real-vs-fake detection, fakes are generated
class-conditionally. For each digit class $d$ we sample independent
latent vectors and generate enough fake images so that, for every split
$s \in \{\mathrm{train}, \mathrm{eval}, \mathrm{test}\}$,
\[
|\mathcal{F}_{s,d}| = |\mathcal{R}_{s,d}|.
\]
The resulting binary datasets are
\[
\mathcal{D}_{s} =
\{(x,0): x \in \mathcal{R}_{s}\}
\cup
\{(x,1): x \in \mathcal{F}_{s}\},
\]
where label $0$ denotes real and label $1$ denotes fake. Both real and
fake images are stored on disk as 8-bit PNGs and reloaded before
training, so any sub-pixel float-precision differences from the
generator are removed before either class enters the pipeline.
Abstractions are applied \emph{after} the split is constructed and
identically to both classes (Sec.~\ref{app:mnist_abstractions}).

\subsubsection{Image-Space Classifiers}
\label{app:mnist_classifiers}

The main MNIST experiments use three binary classifiers: ResNet-Small,
DenseNet-Small, and Mini-Xception. Each takes a single-channel image as
input and outputs one sigmoid probability $h(x) \in [0,1]$, interpreted
as $P(x \text{ is fake})$. Models are trained from scratch separately
for each abstraction level. Architectures are summarized in
Table~\ref{tab:mnist_classifier_architectures}.

\begin{table}[h]
\centering
\small
\begin{tabular}{l p{0.62\linewidth} r}
\toprule
Model & Architecture & Parameters \\
\midrule
ResNet-Small &
$3{\times}3$ convolution stem with $16$ channels, followed by three
residual stages: two $16$-channel residual blocks, two $32$-channel
residual blocks with stride-$2$ downsampling in the first block, and
two $64$-channel residual blocks with stride-$2$ downsampling in the
first block. Each residual block uses two $3{\times}3$ convolutions
with batch normalization and ReLU, plus an identity or $1{\times}1$
projection shortcut. Final adaptive average pooling and a linear
sigmoid head. &
$174{,}945$ \\
\midrule

DenseNet-Small &
DenseNet-BC style network with growth rate $k=12$, initial
$3{\times}3$ convolution to $24$ channels, three dense blocks with
four dense layers per block, and transition layers between blocks.
Each dense layer uses BN--ReLU--$1{\times}1$ convolution followed by
BN--ReLU--$3{\times}3$ convolution, then concatenates features with
the input. Final batch normalization, adaptive average pooling, and a
linear sigmoid head. &
$101{,}479$ \\
\midrule

Mini-Xception &
Compact Xception-style network. Entry block: $1\to 8$ channels via a
$3{\times}3$ convolution, batch normalization, and ReLU. Three
residual depthwise-separable blocks with channel sizes $16$, $32$,
and $64$; the first two downsample by a factor of $2$. Final adaptive
average pooling and a linear sigmoid head. &
$13{,}089$ \\
\bottomrule
\end{tabular}
\caption{Binary classifier architectures used for the MNIST
real-vs-fake abstraction experiments.}
\label{tab:mnist_classifier_architectures}
\end{table}

\subsubsection{Classifier Optimization}
\label{app:mnist_optim}

All image-space classifiers are trained with binary cross-entropy:
\[
\mathcal{L}(h) =
-\frac{1}{|\mathcal{D}_{\mathrm{train}}|}
\sum_{(x,y)\in\mathcal{D}_{\mathrm{train}}}
\bigl[y\log h(x) + (1-y)\log(1-h(x))\bigr].
\]
We use Adam (learning rate $10^{-3}$, default $\beta$), batch size $64$,
random seed $42$, and train for $25$ epochs. A step learning-rate
scheduler is applied per epoch with multiplicative decay factor
$\gamma=0.7$, so the learning rate at epoch $t$ is
$\eta_t = 10^{-3}\cdot 0.7^{t-1}$ (this brings the LR to roughly
$1.3\times 10^{-7}$ by the end of training, effectively freezing the
network in the last few epochs).
Inputs are normalized using the canonical MNIST mean and standard
deviation, $(0.1307, 0.3081)$, after applying the abstraction operator.
At test time, predictions are thresholded at $0.5$, and we report the
overall test accuracy as well as the per-class (real vs.\ fake)
accuracy.

\subsubsection{Abstraction Operators}
\label{app:mnist_abstractions}

Each abstraction operator $f$ is applied to both real and fake images
before classifier training and evaluation. Stochastic operators (Gaussian
noise, uniform noise, salt-and-pepper) draw an independent noise sample
per image using a deterministic per-split RNG (seeded by $42$), so that
every clean image has a single corrupted twin and the procedure is
exactly reproducible.

\paragraph{PNG-roundtrip simulation.}
After applying any abstraction operator $f$, both the clean and
corrupted images are quantized back to 8-bit precision before
normalization:
\[
\tilde{x} = \frac{\mathrm{round}(255 \cdot \mathrm{clip}(x, 0, 1))}{255}.
\]
This step is applied identically to the real branch (which is already
8-bit on disk so the operation is a no-op there) and to the corrupted
branch. Without it, sub-pixel float-precision differences introduced by
$f$ would survive into the network input even though they could never be
observed in any actual saved PNG; we treat such artifacts as
``ghost signals'' that should not count as detectable.

The full preprocessing pipeline is therefore:
\[
x \xrightarrow{f} x_f \xrightarrow{\mathrm{8\text{-}bit\ round}}
\tilde{x}_f \xrightarrow{\mathrm{normalize}(\mu, \sigma)} \hat{x}_f,
\]
where $(\mu,\sigma) = (0.1307,\,0.3081)$ is the canonical MNIST
normalization.

\paragraph{Operator definitions.}
The individual abstraction operators used in the experiments are
defined below.

\begin{itemize}
\item \textbf{Identity:} $f(x)=x$.
\item \textbf{Gaussian blur:} 2-D Gaussian convolution with bandwidth
  $\sigma$. The kernel size is $\max\!\bigl(3,\,
  (\lfloor 6\sigma+0.5\rfloor)\,|\,1\bigr)$ to capture $\geq 99\%$ of
  the kernel mass and stay odd.
\item \textbf{Gaussian noise:} $f(x)=\mathrm{clip}(x+\varepsilon,0,1)$
  with $\varepsilon\sim\mathcal{N}(0,\sigma^2)$ drawn independently
  per pixel.
\item \textbf{Uniform noise:} $f(x)=\mathrm{clip}(x+\varepsilon,0,1)$
  with $\varepsilon\sim\mathrm{Unif}(-a,a)$ drawn independently per
  pixel.
\item \textbf{Classic Salt-and-pepper noise:} a fraction $p$ of pixels (chosen
  by sampling $\lfloor 784\,p\rfloor$ pixel indices without
  replacement) is set to $0$ or $1$ with equal probability; remaining
  pixels are unchanged.
  \item \textbf{Soft salt-and-pepper noise:} a fraction $p$ of pixels
  (chosen by sampling $\lfloor 784\,p\rfloor$ pixel indices without
  replacement) is perturbed by a fixed magnitude rather than forced to
  black or white. For each chosen pixel, we add either $+s$ or $-s$
  with equal probability:
  \[
  f(x)_i = \begin{cases}
    \mathrm{clip}(x_i + s, 0, 1) & \text{w.p. } p/2,\\
    \mathrm{clip}(x_i - s, 0, 1) & \text{w.p. } p/2,\\
    x_i & \text{otherwise.}
  \end{cases}
  \]
  This is distinct from uniform noise, which adds a continuously-drawn
  perturbation to \emph{every} pixel, and from classic salt-and-pepper,
  which forces selected pixels to extreme values regardless of the
  pixel's original content.
\item \textbf{Quantization:} pixel values are reduced to $L$ discrete
  levels,
  $f(x)=\mathrm{round}\!\bigl(x(L-1)\bigr)/(L-1)$.
\item \textbf{Brightness shift:} $f(x)=\mathrm{clip}(x+b,0,1)$ for
  signed shift $b\in\mathbb{R}$.
\item \textbf{Sharpening (unsharp masking):} $f(x) = \mathrm{clip}\!\bigl(x + \alpha\,(x - g_{\sigma=1}(x)),\,0,1\bigr)$,
  where $g_{\sigma=1}$ is a 2-D Gaussian blur with $\sigma=1$ (yielding
  a $7{\times}7$ kernel via the same kernel-size rule above) and
  $\alpha\geq 0$ is the sharpening amount. $\alpha=0$ is identity;
  $\alpha=1$ is moderate; $\alpha\geq 2$ produces noticeable edge
  enhancement.
\item \textbf{Downsampling:} the image is bilinearly resized from
  $28{\times}28$ to $N{\times}N$ and then bilinearly upsampled back
  to $28{\times}28$ — i.e., a low-pass filter that keeps only the
  information representable at resolution $N$.
\item \textbf{Center crop:} the central $N{\times}N$ patch is retained,
  then bilinearly upsampled back to $28{\times}28$. This zooms into the
  central content rather than blacking out the periphery, so the
  classifier sees the same input shape with all available signal coming
  from the cropped region.
\item \textbf{PCA:} each image is flattened to $784$ dimensions,
  standardized using training-set statistics, projected onto the top
  $k$ principal components, and classified directly in PCA space
  (Sec.~\ref{app:mnist_pca}).
\end{itemize}

For mixed abstractions, operators are applied sequentially in the
order written; for example, ``blur $2.0$ + quant $4$ + gauss $0.1$''
denotes
\[
f(x) = f_{\mathrm{noise}}\!\bigl(f_{\mathrm{quant}}\!\bigl(f_{\mathrm{blur}}(x)\bigr)\bigr).
\]

\subsubsection{The 25 Abstraction Configurations}
\label{app:mnist_25_configs}

\begin{figure*}[t!]
    \centering
    \includegraphics[width=1.0\textwidth]{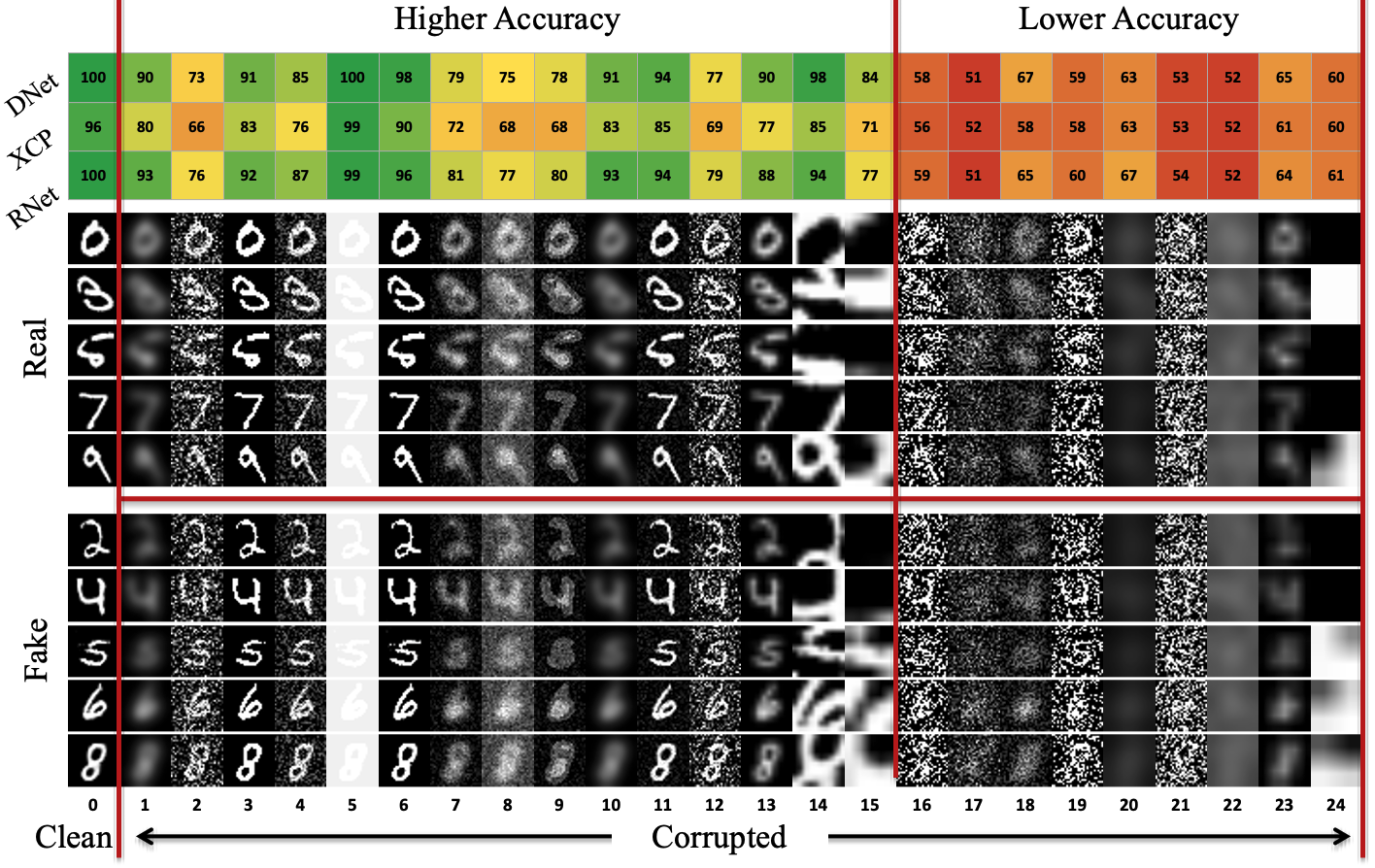}
    \caption{
MNIST real-vs-fake detection under image abstractions. Columns show clean and corrupted versions of matched real and ACGAN-generated digits, while the heatmap reports test accuracy for DenseNet-Small (DNet), Mini-Xception (XCP),
and ResNet-Small (RNet). Real and fake images remain separable through many abstractions and approach chance only when the digit content is heavily destroyed. Please see Table - ~\ref{tab:mnist_25_configs} for corruption configurations.
}
\label{fig:mnist_results_full}
\end{figure*}

Table~\ref{tab:mnist_25_configs} lists the $25$ configurations
referenced by index in the corresponding figure. Configurations~$0$--$15$
appear in the ``higher accuracy'' (still detectable) panel; $16$--$24$
appear in the ``lower accuracy'' (close to indistinguishable) panel.

\begin{table}[h]
\centering
\scriptsize
\begin{tabular}{r p{4.55cm}}
\toprule
\# & Operation \\
\midrule
0  & clean \\
1  & blur $\sigma{=}3.0$ \\
2  & gauss $\sigma{=}0.5$ \\
3  & quant $L{=}2$ \\
4  & unif $a{=}0.5$ \\
5  & brightness $b{=}+0.95$ \\
6  & sharpen $\alpha{=}5.0$ \\
7  & blur $\sigma{=}2.0$ + gauss $\sigma{=}0.1$ \\
8  & bright $b{=}0.3$ + blur $\sigma{=}2.0$ + gauss $\sigma{=}0.1$ \\
9  & blur $\sigma{=}2.0$ + quant $L{=}4$ + gauss $\sigma{=}0.1$ \\
10 & sharp $\alpha{=}3$ + blur $\sigma{=}3$ \\
11 & quant $L{=}4$ + gauss $\sigma{=}0.1$ \\
12 & quant $L{=}2$ + gauss $\sigma{=}0.5$ \\
13 & downsample $N{=}10$ \\
14 & crop $N{=}10$ \\
15 & crop $N{=}5$ \\
16 & unif $a{=}1.0$ + quant $L{=}2$ \\
17 & blur $\sigma{=}5.0$ + gauss $\sigma{=}0.3$ \\
18 & blur $\sigma{=}3.0$ + unif $a{=}0.3$ \\
19 & gauss $\sigma{=}0.7$ + quant $L{=}2$ \\
20 & blur $\sigma{=}7.5$ \\
21 & unif $a{=}1.5$ \\
22 & unif $a{=}1.5$ + blur $\sigma{=}5.0$ \\
23 & downsample $N{=}5$ \\
24 & crop $N{=}2$ \\
\bottomrule
\end{tabular}
\caption{The 25 abstraction configurations used for the main MNIST experiment.}
\label{tab:mnist_25_configs}
\end{table}

\subsubsection{PCA-Space Classifiers}
\label{app:mnist_pca}

For PCA experiments, real and fake images are flattened into
$784$-dimensional vectors. Standardization is fit only on the training
split and then applied to the eval and test splits. PCA is also fit
only on the standardized training data. We evaluate multiple reduced
dimensions,
\[
k \in \{2,3,4,5,6,8,10,20,50,100,200,400,600\}.
\]
The PCA classifiers are trained with Adam (learning rate $10^{-3}$),
batch size $256$, seed $42$, and $25$ epochs. The main PCA model is a
one-dimensional residual MLP: a linear projection to $256$ hidden
units, three residual fully-connected blocks, and a final linear
sigmoid head.

\subsubsection{Controls and Implementation Notes}
\label{app:mnist_controls}

\paragraph{Real-vs-real negative control.}
We additionally train the same classifiers on a random $50{:}50$ split
of $\mathcal{R}$ (i.e., real-vs-real, with no fake involved). Test
accuracy stays at chance ($\approx 50\%$) across all three architectures
and all abstraction operators, confirming that the real-vs-fake signal
the detectors learn cannot be attributed to incidental split identity
or class-imbalance shortcuts.

\paragraph{Digit-balanced fake splits.}
For real-vs-fake experiments, fake splits are digit-matched to the real
splits: $|\mathcal{F}_{s,d}| = |\mathcal{R}_{s,d}|$ for every split
$s$ and digit class $d$. This prevents the classifier from exploiting
any digit-frequency imbalance between real and fake.

\paragraph{Reporting.}
All reported test accuracies are computed on the held-out test split
$\mathcal{R}_{\mathrm{test}} \cup \mathcal{F}_{\mathrm{test}}$, which is
used neither for training nor for model selection. We report the
final-epoch test accuracy after the $25$-epoch schedule.

\paragraph{Reproducibility.}
All experiments use a fixed random seed of $42$ for the abstraction
RNG, model initialization, and the data shuffle. Models are trained on
single NVIDIA GPUs of the institutional cluster; the full sweep
of $\bigl|\mathcal{H}\bigr|\times$ (number of abstractions) jobs
completes in under a few GPU-hours per architecture. All training,
evaluation, and plotting code is released alongside the paper.

%%%%%%%%%%%%%%%%%%%%% BEGIN - CelebA description %%%%%%%%%%%%%%%%%%%%%%%%

\subsection{CelebA}\label{app:celeba}

\textbf{Dataset.}
202,599 real faces from CelebA paired with 202,599 DCGAN-generated faces
(64$\times$64 RGB). Training: 323,158 images; test: 81,040 images.

\textbf{Ablations applied to both real and generated images:}
\begin{itemize}
  \item \textbf{Downsampling}: bilinear resize to $\{32, 16, 8, 4, 2, 1\}^2$ pixels.
  \item \textbf{Gaussian blur}: $\sigma \in \{6, 8, 10, 15, 20, 30\}$; image stays 64$\times$64.
  \item \textbf{Center crop}: sizes $\{10, 4, 2, 1\}^2$ pixels extracted from the image center.
\end{itemize}

\textbf{Training hyperparameters.}
50 epochs, batch size 64.
Baseline CNN uses Adadelta (lr $= 1.0$); all other architectures use Adam (lr $= 10^{-3}$).
No weight decay.

\textbf{Architectures} (all adapted for 3-channel 64$\times$64 RGB input):

\begin{table}[h]
\centering
\small
\begin{tabular}{|l|l|r|p{7cm}|}
\hline
\textbf{Name} & \textbf{Optimizer} & \textbf{Params} & \textbf{Description} \\
\hline
Baseline CNN    & Adadelta 1.0   & 7,392,449 & The model uses two convolutional layers, mapping $3$ channels to $32$ and
then $32$ to $64$, followed by max pooling and dropout with probability
$0.25$. The flattened features pass through a $128$-unit fully connected
layer, dropout with probability $0.5$, and a final one-unit output layer.
 \\
\hline
CNN+BatchNorm   & Adam $10^{-3}$ & 7,392,641 & Baseline with BatchNorm after each conv; dropout removed. \\
\hline
AC-GAN Disc.    & Adam $10^{-3}$ & 99,937    & 4 stride-2 conv layers, LeakyReLU(0.2), BatchNorm, Dropout(0.25). Mirrors AC-GAN discriminator. \\
\hline
ResNet-Small    & Adam $10^{-3}$ & 175,233   & 3$\times$3 stem; 3 residual blocks (16$\to$32$\to$64 ch, stride-2); AdaptiveAvgPool; FC(64$\to$1). \\
\hline
Constrained CNN & Adam $10^{-3}$ & 189,505   & Zero-sum constrained first conv (5$\times$5); 2 conv layers; AdaptiveAvgPool(4); FC(1024$\to$128$\to$1). \\
\hline
DenseNet-Small  & Adam $10^{-3}$ & 101,911   & 3 dense blocks, growth rate $k=12$; AdaptiveAvgPool. \\
\hline
Mini-Xception   & Adam $10^{-3}$ & 13,233    & Depthwise separable convolutions; AdaptiveAvgPool. \\
\hline
SRNet           & Adam $10^{-3}$ & 362,593   & Steganalysis residual network; no early pooling to preserve high-frequency artifacts. \\
\hline
\end{tabular}
\caption{Architectures used in the CelebA experiments.}
\label{tab:celeba_classifiers}
\end{table}

\begin{figure}[htbp]
    \centering
    \includegraphics[width=0.8\textwidth]{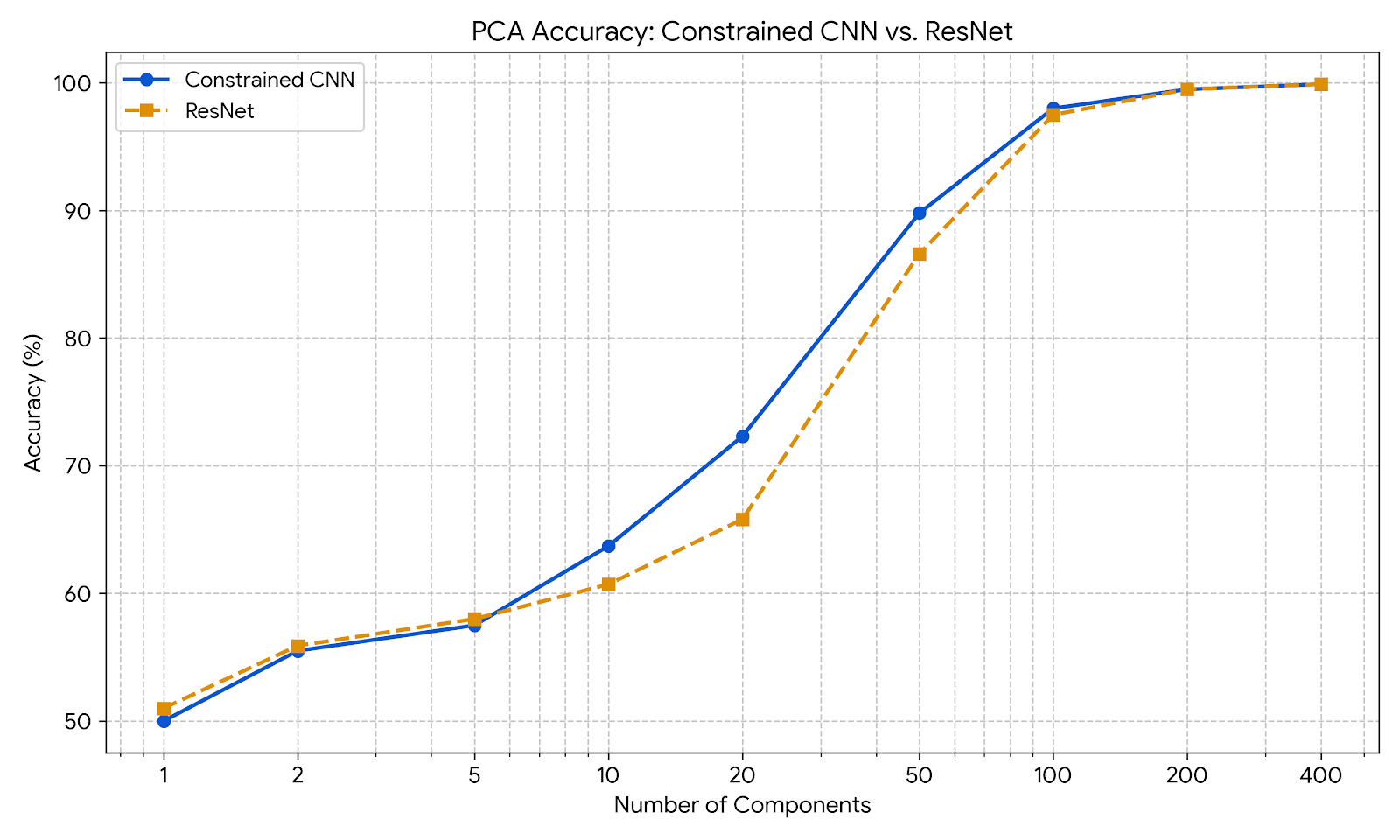}
    \caption{PCA results for CelebA/DcGAN configuration}
    \label{fig:pca_results_celeba}
\end{figure}

We experiment with CelebA~\cite{liu2015faceattributes}, a slightly less complex version of FFHQ. Specifically, we use the cropped and aligned version which has $\sim 200$k images at $64 \times 64$ resolution. This dataset, $\mathcal{R}$, is split between $\mathcal{R}_{train}$ and $\mathcal{R}_{test}$ having $160$k and $40$k images respectively. We train DcGAN~\cite{radford-iclr16} on $\mathcal{R}_{train}$ for 10 epochs, whose latent input just consists of a single noise $z \sim \mathcal{N} (0,1)$. Post training, we generate fake image datasets $\mathcal{F}_{train}$ and $\mathcal{F}_{test}$ having the same number of images as the real counterpart. In this case, out family of classifiers $\mathcal{H}$ includes ConstrainedCNN~\cite{constrained_cnn}, DenseNet and SRNet. Similar to above, the classifiers are trained on $\mathcal{D}_{train}$ for 25 epochs and evaluated on $\mathcal{D}_{test}$. Furthermore, following the advice from \cite{patchforensics}, we process the real and fake images to make sure that there are no differences due to formatting issues (e.g., one being saved in PNG and the other in JPEG). This way, the only differences that get discovered between $\mathcal{R}$ and $\mathcal{F}$ by $\mathcal{H}$ can safely be attributed to the errors of $G$.

\begin{figure*}[t!]
    \centering
    \includegraphics[width=1.0\textwidth]{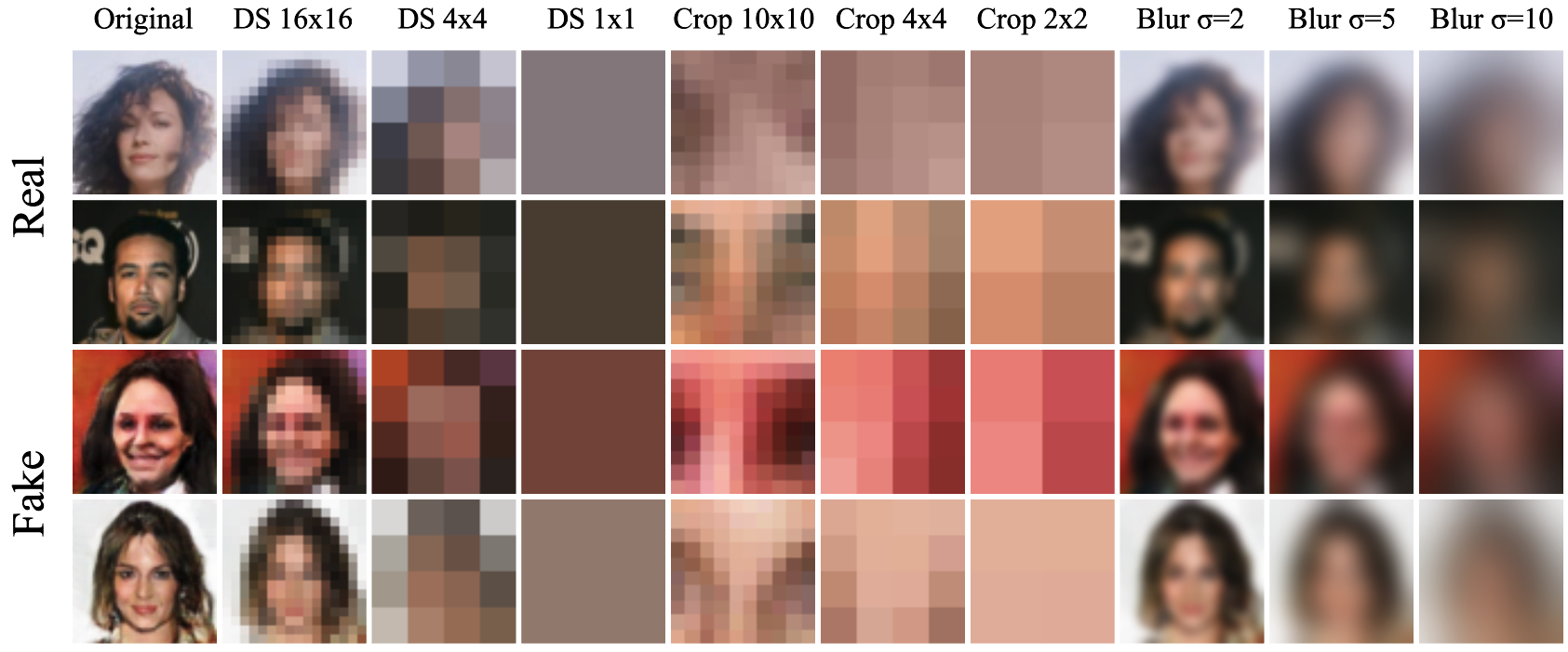}
    \caption{
CelebA real and DCGAN-generated faces under spatial and blur abstractions.
}
    \label{fig:celeba_results}
\end{figure*}

We consider three types of abstractions: (i) downsizing, (ii) blurring, (iii) cropping. The results are shown in Fig.~\ref{fig:celeba_results} and Table~\ref{tab:celeba_main}.
. For each abstraction type, the strength of the operation (e.g., the amount of blur, crop) increases as we move from left to right. Generally speaking, the accuracy of the classifiers decreases along that trend. However, similar to MNIST, what is interesting is how often it stays above $>50\%$. For example, even after downsampling $\mathcal{R}$ and $\mathcal{F}$ from $64 \times 64$ to $4 \times 4$, one of the $h$ can still distinguish the two categories $90\%$ of the time. A similarly surprising example is with blurring with $\sigma=30$. That is so much blurring that it is hard to know if there was even a face in these images. So, it is safe to assume that all semantic anomalies, if they were present in any particular fake image, would have been removed. In spite of this, they are easily separable by all classifiers. We also show the results on PCA for two classifiers in Fig. ~\ref{fig:pca_results_celeba}. We observe that one can compress the data more than $200$ times ($64 \times 64 \times 3 \rightarrow 50$) and still have a classifier separate $\sim90\%$ of data points. The only cases where all the classifiers perform at around chance performance ($50-55\%$) are either if one downsizes the real/fake images to a single pixel or compresses the original signal more than $6000$ times using PCA. 

%%%%%%%%%%%%%%%%%%%%% BEGIN - CelebA results %%%%%%%%%%%%%%%%%%%%%%%%

\begin{table*}[]
\resizebox{\textwidth}{!}{
\begin{tabular}{lcccccccccccc}
\toprule
\multirow{2}{*}{\textbf{Model}} 
& \multicolumn{4}{c}{Downsizing} 
& \multicolumn{4}{c}{Cropping} 
& \multicolumn{4}{c}{Blurring} 

\\
\cmidrule(lr){2-5} \cmidrule(lr){6-9} \cmidrule(lr){10-13}  
& $16 \times 16$ & $4 \times 4$ & $2 \times 2$ & $1 \times 1$ & $10 \times 10$ & $4 \times 4$ & $2 \times 2$ & $1 \times 1$ & $\sigma=6$ & $\sigma=10$ & $\sigma=20$ & $\sigma=30$ \\
\midrule
Constrained CNN 
& 99.7\% & 90.2\%
& 68.9\% & 53.9\%
& 100.0\% & 99.9\%   
& 94.6\% & 57.1\% 
& 97.3\% & 91.7\%   
& 93.4\% & 96.2\%      
 \\
DenseNet
& 99.5\% & 89.0\% 
& 68.2\% & 54.0\% 
& 100.0\% & 100.0\%   
& 94.6\% & 57.1\% 
& 98.2\% & 91.0\%   
& 90.6\% & 91.2\%      

\\

SRNet 
& 99.8\% & 90.5\%   
& 68.4\% & 54.1\%     
& 100.0\% & 99.9\%   
& 94.5\% & 57.1\%      
& 96.4\% & 84.1\%   
& 83.6\% & 86.5\%      

\\

\bottomrule
\end{tabular}
}
\caption{CelebA real-vs-fake test accuracy under downsampling, cropping, and blurring
abstractions. Each entry reports binary classification accuracy on held-out
real and DCGAN-generated test images. Detection remains well above chance for
most abstractions and approaches chance only under extreme spatial collapse,
such as $1\times1$ downsampling or cropping.}
\label{tab:celeba_main}
\end{table*}

%%%%%%%%%%%%%%%%%%%%% END - CelebA results %%%%%%%%%%%%%%%%%%%%%%%%

%%%%%%%%%%%%%%%%%%%%% BEGIN - FFHQ Description %%%%%%%%%%%%%%%%%%%%%%%%

\subsection{FFHQ}\label{app:ffhq}

\textbf{Dataset.}
70,000 real FFHQ faces paired with 70,000 StyleGAN2/VD VAE/LDM-generated faces (256$\times$256).
Training: 112,000 images; test: 28,000 images.
All models are trained for 50 epochs, batch size 64, weight decay $10^{-4}$.
CNN classifiers use Adam (lr $= 10^{-3}$); ViT uses Adam (lr $= 10^{-4}$).

Two families of inputs are evaluated. The first operates on \textbf{derived representations}
rather than raw pixels:
\begin{itemize}
  \item \textbf{Canny edge maps}: grayscale binary edge images computed with \texttt{cv2.Canny()}.
  \item \textbf{Face segmentation maps}: per-pixel class labels (0--18) produced by a pretrained
        BiSeNet (ResNet-18 backbone) fine-tuned on CelebAMask-HQ.
        Raw integer label maps are used as classifier input.
\end{itemize}
The second operates directly on \textbf{raw RGB pixels}, with the following ablations:
\begin{itemize}
  \item \textbf{Center crop}: sizes $\{64, 16, 4, 1\}^2$ pixels.
  \item \textbf{Downsampling}: bilinear resize to $\{64, 16, 4, 1\}^2$ pixels.
  \item \textbf{Gaussian blur}: $\sigma \in \{5, 10, 20, 40\}$; image stays 256$\times$256.
\end{itemize}

The same three architectures are used across both input families.
For derived representations the input is 1-channel grayscale; for raw RGB the first
convolution is adapted to accept 3 channels.

\begin{table}[h]
\centering
\small
\begin{tabular}{|l|l|r|r|p{5.5cm}|}
\hline
\textbf{Name} & \textbf{Optimizer} & \textbf{Params (1ch)} & \textbf{Params (3ch)} & \textbf{Description} \\
\hline
ResNet  & Adam $10^{-3}$ & 174,385 & 174,817 & Stem Conv+BN+ReLU; 3 residual stride-2 layers (16$\to$32$\to$64 ch); AdaptiveAvgPool; FC(64$\to$1). \\
DenseNet & Adam $10^{-3}$ & 133,481 & 134,057 & Stem Conv(stride=2)+MaxPool; 3 DenseBlocks (growth rate 16, 4 layers each); AdaptiveAvgPool; FC$\to$Sigmoid. \\
ViT     & Adam $10^{-4}$ & 12,874,241 & 12,875,265 & patch\_size=16, dim=512, depth=6, heads=8, mlp\_dim=1024, dropout=0.1. \\
\hline
\end{tabular}
\caption{Architectures used in the FFHQ experiments. Parameter counts are given for
1-channel (derived representations) and 3-channel (raw RGB) inputs.}
\label{tab:ffhq_classifiers}
\end{table}

\paragraph{Per-face-part dataset construction.}
In addition to the global classifiers above, we train one binary
classifier per face part on tight crops localized to that region.
Crops are produced from the same 70k real FFHQ + 70k StyleGAN2 images
in two passes: (i) a one-time bounding-box localization pass, and
(ii) a fixed-size cropping pass driven by per-part templates derived
from the localization statistics.

\textbf{Bounding-box localization.}
We start from CelebAMask-HQ-style segmentation maps (one class label
per pixel, 19 classes covering eyes, brows, nose, mouth, ears, hair,
etc.), produced by the same BiSeNet parser used above for the
segmentation-map experiments. For each of 14 target parts (eye, nose,
mouth, mouth\_inner, left/right\_eye, brows, left/right\_brow,
left/right\_ear, eye\_g (glasses), neck, hair) we map the part to the
relevant class IDs (e.g.\ \emph{eye} = brows $\cup$ eyes,
\emph{nose} = class 10, \emph{left\_eye} = class 4,
\emph{eye\_g} = class 6) and extract the pixel mask of those classes.
A tight axis-aligned bounding box is then computed per image per part
with two robustness adjustments:
\begin{itemize}
  \item \textbf{Percentile-trimmed extents.} Instead of the full
        min/max of the mask, we use the 1st and 99th percentile on
        each axis (i.e.\ keep the central $98\%$ of pixels). This
        suppresses parser hallucinations that occasionally place a
        few stray pixels far from the true region.
  \item \textbf{Largest connected component.} For laterally paired
        parts (\emph{left/right\_eye}, \emph{left/right\_brow},
        \emph{left/right\_ear}) we additionally restrict the mask to
        the largest 4-connected component before computing the box,
        which discards scattered mis-classified pixels on the wrong
        side of the face.
\end{itemize}
A part is marked \emph{excluded} when glasses (class 6) overlap the
eyes (so eye crops would contain frames rather than skin), and
\emph{invalid} when fewer than 20 pixels of the target class are
present. The resulting boxes are written to JSON per part, alongside
a status flag, so downstream stages can read only the \texttt{ok}
boxes without recomputing segmentation.

\textbf{Per-part fixed crop templates.}
Per-image bounding boxes vary in size and aspect ratio, so we
collapse them to one fixed crop size per part. For each part we
collect the widths and heights of all \texttt{ok} boxes across both
classes (real and fake) and pick the 98th percentile of each,
rounded up to the nearest multiple of 8 (template
$W_p \times H_p$). The choice of P98 is motivated empirically: the
width distributions exhibit a knee around the 98th percentile, after
which boxes grow primarily because of parser noise or extreme poses
rather than because the part is genuinely larger. Rounding to a
multiple of 8 makes the crop dimensions friendly for both CNN strides
and ViT patch grids without introducing artificial padding pixels.

\textbf{Cropping pass.}
Given the per-part template $(W_p, H_p)$ and the per-image bounding
box $(x_0, y_0, x_1, y_1)$, we compute the bounding-box center
$(c_x, c_y)$ and carve out a window of size $W_p \times H_p$ centered
on $(c_x, c_y)$. If the window would fall outside the image we shift
it inward (clamp), which preserves the template size at the cost of a
small location offset for boundary cases (typical for ear crops near
the image edge). No resampling, padding, or interpolation is applied:
all crops are exact pixel sub-windows of the source image, so the
real-vs-fake comparison cannot be biased by an asymmetric resampling
pipeline. Crops are written as PNG with no metadata, again matching
real and fake byte-for-byte except in pixel content.

\textbf{Sanity checks.}
Before training we verify that (a) for every part the cardinality of
real and fake \texttt{ok} crops differs by less than $2\%$, (b) the
file-size, image-resolution, and per-channel mean/std distributions
of real and fake crops are statistically indistinguishable
(logistic-regression baseline accuracy $\le 55\%$ on each, all near
chance), and (c) the train/val/test split (70/15/15, deterministic
md5 hash of \texttt{image\_id}) places every \texttt{image\_id} in
exactly one split, so a given source identity never appears across
splits even when the same identity is cropped for multiple parts.
After per-split class balancing (subsample the larger class to match
the smaller, deterministic seed) the real and fake counts are exactly
equal in every split for every part.

\textbf{Per-part classifiers.}
The same three architectures from above (ResNet, DenseNet, ViT) are
trained once per (part, architecture) pair on the 3-channel RGB
crop. CNN architectures keep the $10^{-3}$ Adam recipe used for the
global FFHQ classifiers; ViT switches to a tuned recipe for these
small per-part crops, in which patch size is reduced from 16 to 4
(yielding $\sim 16\times$ as many tokens given the smaller spatial
extent), AdamW with weight decay $0.05$ replaces Adam, the learning
rate is set to $5\times 10^{-4}$, and a 5-epoch linear warm-up is
followed by a cosine schedule. This change is necessary because the
default Adam $10^{-3}$ + StepLR recipe leaves ViT at 63--72\% test
accuracy on these crops; the smaller-patch / AdamW recipe closes the
gap to within 0.1--0.2\% of the CNN baselines.

%%%%%%%%%%%%%%% BEGIN - FFHQ results %%%%%%%%%%%%%%%
%%%%% StyleGAN %%%%%%%%%%%%%%%

\begin{figure*}[t!]
    \centering
    \includegraphics[width=1.0\textwidth]{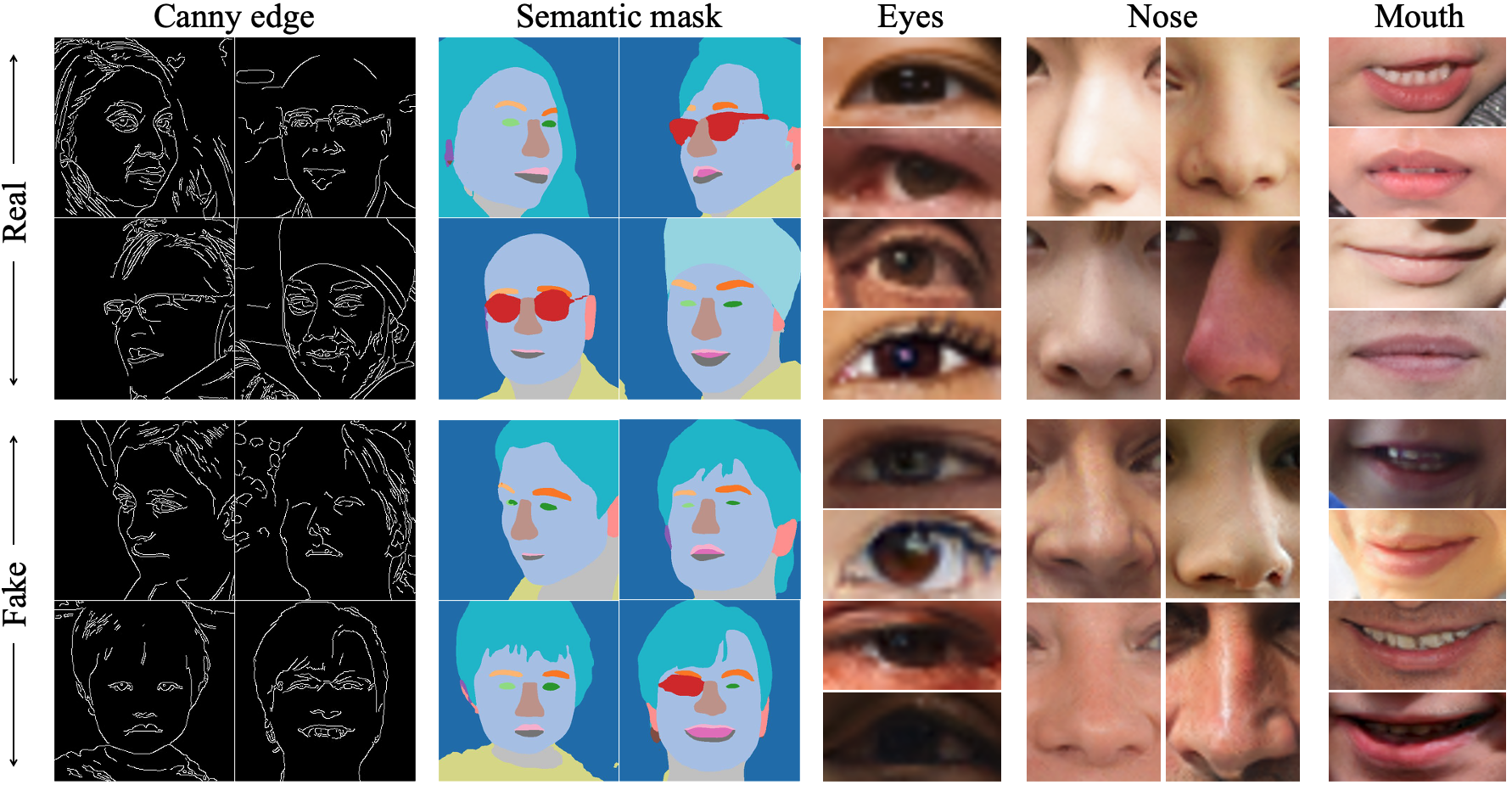}
    \caption{FFHQ real and StyleGAN2-generated faces under semantic and local-part
abstractions. }
    \label{fig:ffhq_v2_stylegan}
\end{figure*}

\begin{figure*}[t!]
    \centering
    \includegraphics[width=1.0\textwidth]{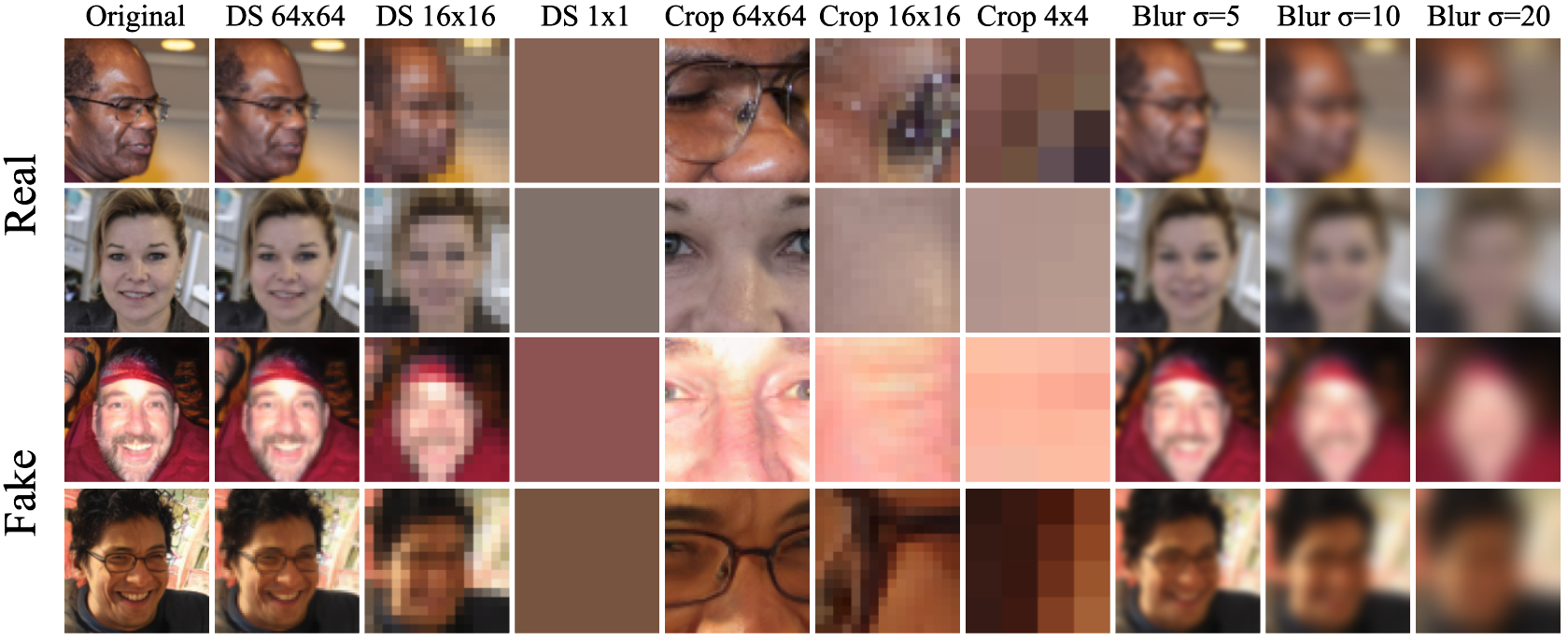}
    \caption{FFHQ real and StyleGAN2-generated faces under spatial and blur abstractions.}
    \label{fig:ffhq_results_stylegan}
\end{figure*}

\begin{table*}[]
\resizebox{\textwidth}{!}{
\begin{tabular}{lcccccccccccccc}
\toprule
\multirow{2}{*}{\textbf{Model}}
& \multicolumn{4}{c}{Downsizing}
& \multicolumn{4}{c}{Cropping}
& \multicolumn{4}{c}{Blurring}
\\
\cmidrule(lr){2-5} \cmidrule(lr){6-9} \cmidrule(lr){10-13} \cmidrule(lr){14-14} \cmidrule(lr){15-15}
& $64{\times}64$ & $16{\times}16$ & $4{\times}4$ & $1{\times}1$
& $64{\times}64$ & $16{\times}16$ & $4{\times}4$ & $1{\times}1$
& $\sigma{=}5$ & $\sigma{=}10$ & $\sigma{=}20$ & $\sigma{=}40$ \\
\midrule
ResNet
& 99.1\% & 67.2\% & 54.5\% & 52.6\%
& 100.0\% & 99.7\% & 81.7\% & 54.7\%
& 96.5\% & 76.9\% & 57.0\% & 55.2\% \\
DenseNet
& 99.5\% & 77.6\% & 55.5\% & 52.5\%
& 100.0\% & 99.8\% & 96.3\% & 54.8\%
& 96.7\% & 82.3\% & 66.8\% & 59.6\% \\
ViT
& 99.8\% & 86.2\% & 58.4\% & 51.3\%
& 100.0\% & 99.8\% & 95.1\% & 50.5\%
& 98.0\% & 93.0\% & 74.0\% & 63.0\% \\
\bottomrule
\end{tabular}
}
\caption{FFHQ real-vs-fake test accuracy under downsampling, cropping, and blurring
abstractions. Each entry reports binary classification accuracy on held-out
real and StyleGAN2-generated test images. Detection remains high for many
spatially degraded inputs and approaches chance only under the strongest
abstractions, such as $1\times1$ spatial collapse.}
\label{tab:ffhq_stylegan_detailed}
\end{table*}

\begin{table}[t]
\centering
\resizebox{1.0\textwidth}{!}{%
\begin{tabular}{lcccccccc}
\toprule
& Edge    &  Mask    &  Eye    &  Nose    &  Mouth    &  L-eye    &  R-brow    &  R-ear        \\
\midrule
ResNet   & 82.4\% & 71.1\% & 99.69\% & 99.28\% & 99.85\% & 98.97\% & 98.59\% & 74.97\% \\
DenseNet & 85.9\% & 76.1\% & 99.84\% & 99.90\% & 99.71\% & 99.24\% & 97.19\% & 86.19\% \\
ViT      & 91.7\% & 70.9\% & 99.83\% & 99.78\% & 99.77\% & 98.34\% & 98.88\% & 98.95\% \\
\bottomrule
\end{tabular}
}
\caption{Real-vs-fake test accuracy on FFHQ vs.\ StyleGAN2 across derived
representations (Edge, Mask) and six cropped face parts. }
\label{tab:ffhq_v2_stylegan}
\end{table}

%%%%%%%%%% VD VAE %%%%%%%%%%%%%%%%%%%%%%
%%%%%%%%%% LDM %%%%%%%%%%%%%%%%%%%%%%

%%%%%%%%%%%%%%% END - FFHQ results %%%%%%%%%%%%%%%

%%%%%%%%%%%%%%%%%%%%% BEGIN - GenImage Description %%%%%%%%%%%%%%%%%%%%%%%%

\subsection{GenImage}\label{app:genimage}

\textbf{Dataset.}
We use the ImageNet-based GenImage benchmark ~\cite{zhu2023genimage}, which pairs real
ImageNet photographs (the \texttt{nature} split) with images produced by
several text-to-image generators prompted with the $1{,}000$ ImageNet class
names ($\sim$162 generated images per class per generator).
We evaluate four generators: Midjourney, Wukong, Stable Diffusion v1.4, and
Stable Diffusion v1.5. For every generator we use the
same pool of real images: all $167{,}701$ \texttt{nature} images
($161{,}701$ train + $6{,}000$ val) across all $1{,}000$ classes.

\textbf{Format and resolution matching.}
GenImage's real and generated images differ in native resolution, aspect ratio,
and file format. The generated images are square ($1024{\times}1024$ for
Midjourney, $512{\times}512$ for Wukong and Stable Diffusion) and saved as PNG.
The real ImageNet photographs are JPEGs with arbitrary aspect ratios (median
$\sim 500{\times}375$). To prevent the classifier from exploiting these
formatting differences~\cite{patchforensics}, all images are brought to a
common representation before training.
Resizing the real images straight to a square distorts their aspect ratio.
We checked this by comparing grids of randomly sampled real and generated
images from the same ImageNet class, and the distortion was visible.
We therefore center-crop each real image to a square of side $\min(h, w)$,
which preserves its aspect ratio, and then resize it to $256{\times}256$.
Each generated image, which is already square, is resized directly to
$256{\times}256$. Both resizes use Lanczos resampling, and every image in both
classes is stored as a lossless 8-bit PNG. This single set of processed
real images is shared across all four generators.

\textbf{Splits.}
GenImage's official train and val folders are merged. Each generator provides
more generated images than there are real images, so the larger class is
subsampled to the size of the smaller one. The balanced pool is then split at
random into 80\% train and 20\% test (seed $42$). For Midjourney this yields
$268{,}322$ training and $67{,}080$ test images, split equally between real
and fake.

\textbf{Ablations applied to both real and generated images:}
\begin{itemize}
  \item \textbf{Downsampling}: bilinear resize to $\{64, 16, 4, 1\}^2$ pixels,
        then bilinear upsampling back to $256{\times}256$.
  \item \textbf{Gaussian blur}: $\sigma \in \{5, 10, 20, 40\}$ at $256{\times}256$.
  \item \textbf{Center crop}: sizes $\{64, 16, 4, 1\}^2$ pixels taken from the
        image center, then bilinearly upsampled back to $256{\times}256$.
  \item \textbf{JPEG compression}: re-encoding at quality $q \in \{75, 50, 25, 5\}$
        and decoding again before the image is passed to the network.

\end{itemize}
Every ablation is applied on the fly and identically to both classes, and the
output always has the same $256{\times}256{\times}3$ input shape. A separate
network is trained from scratch for each (generator, ablation, strength)
combination.

\textbf{Training hyperparameters.}
We use the 3-channel DenseNet from the FFHQ experiments
(Table~\ref{tab:ffhq_classifiers}; $134{,}057$ parameters). It is trained with
binary cross-entropy, Adam (lr $= 10^{-3}$, weight decay $10^{-4}$), and a
ReduceLROnPlateau scheduler (factor $0.5$, patience $5$) for 50 epochs with
batch size 64. Inputs are normalized to $[-1, 1]$.
Because the full data pool ($\sim$36\,GB of PNGs) exceeds what the cluster's
network file system can serve efficiently as individual files, the images are
stored in tar shards and loaded into memory once per job. This changes only
the I/O and has no effect on the data.

\textbf{Real-vs-real control.}
For each generator's setup we also train the same DenseNet on a random 50/50
split of the real images, with no generated images involved. This checks
whether the preprocessing pipeline itself gives the classifier an incidental
cue.

%%%%%%%%%%%%%%%%%%%%% END - GenImage Description %%%%%%%%%%%%%%%%%%%%%%%%

%%%%%%%%%%%%%%% BEGIN - GenImage results %%%%%%%%%%%%%%%

\begin{figure*}[t!]
    \centering
    \includegraphics[width=1.0\textwidth]{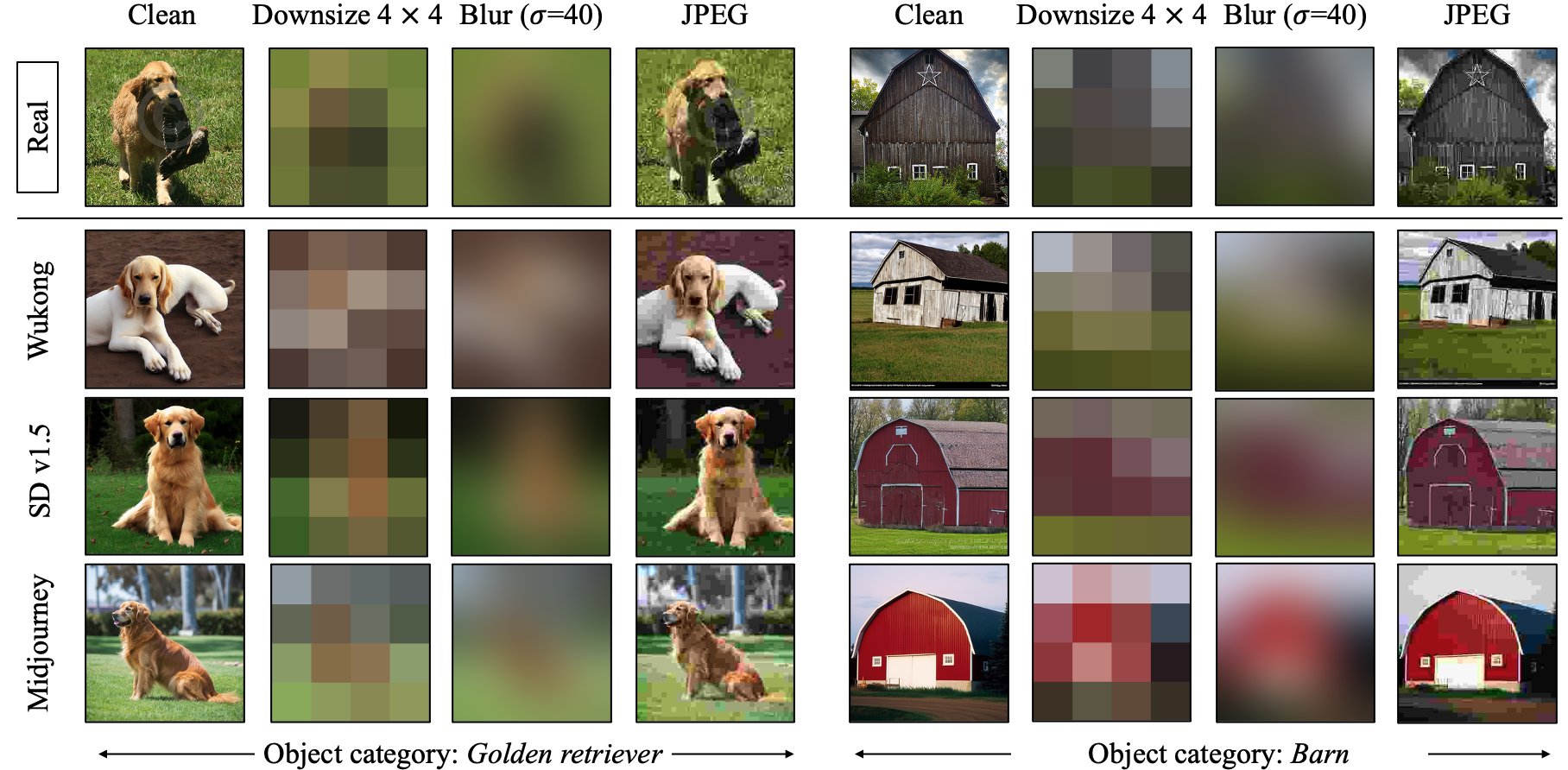}
    \caption{Real (top) and fake images (bottom three) from GenImage dataset under different spatial and blur abstractions.}
    \label{fig:genimage_results_v2}
\end{figure*}

\end{document}